\documentclass[preprint,12pt]{elsarticle}
\usepackage{amssymb}
\usepackage{amsmath}
\usepackage{graphicx}
\usepackage{amsmath,amssymb} % define this before the line numbering.
\usepackage{multirow}
\usepackage{adjustbox}
\usepackage{booktabs}
\usepackage{color}
\usepackage{subcaption}
\usepackage{algorithm}
\usepackage{algpseudocode}
\usepackage{epstopdf}
\journal{}

\usepackage{listings}
\usepackage{xcolor}
\usepackage{caption}

\definecolor{codegray}{rgb}{0.5,0.5,0.5}
\definecolor{backcolor}{rgb}{0.95,0.95,0.95}

\lstdefinestyle{pythonstyle}{
    backgroundcolor=\color{backcolor},   
    commentstyle=\color{codegray},
    keywordstyle=\color{blue},
    numberstyle=\tiny\color{codegray},
    stringstyle=\color{red},
    basicstyle=\ttfamily\footnotesize,
    breaklines=true,
    captionpos=b,
    keepspaces=true,
    numbers=left,
    numbersep=5pt,
    showspaces=false,
    showstringspaces=false,
    showtabs=false,
    tabsize=2,
    language=Python
}

\begin{document}

\begin{frontmatter}

% \title{Enhancing Vortex Shedding Prediction with Progressive Graph Neural Networks}

\title{Multi-Stage Graph Neural Networks for Data-Driven Prediction of Natural Convection in Enclosed Cavities}

\author[inst1]{Mohammad Ahangarkiasari\corref{cor1}}
\ead{ahangar100@gmail.com}

\author[inst2]{Hassan Pouraria\corref{cor1}}
\ead{}

\affiliation[inst1]{organization={Department of Electrical and Computer Engineering, Aarhus University},
            % addressline={}, 
            city={Aarhus},
            postcode={8200}, 
            % state={State Two},
           country={Denmark}}

\affiliation[inst2]{organization={Department of Chemical and Materials Engineering, New Mexico State University},
            % addressline={}, 
            city={Las Cruces},
            postcode={NM 88003},
            country={USA}}

\cortext[cor1]{Corresponding author}

\begin{abstract}

Buoyancy-driven heat transfer in closed cavities serves as a canonical testbed for thermal design 
High-fidelity CFD modelling yields accurate thermal field solutions, yet its reliance on expert-crafted physics models, fine meshes, and intensive computation limits rapid iteration.
Recent developments in data-driven modeling, especially Graph Neural Networks (GNNs), offer new alternatives for learning thermal-fluid behavior directly from simulation data, particularly on irregular mesh structures. However, conventional GNNs often struggle to capture long-range dependencies in high-resolution graph structures. To overcome this limitation, we propose a novel multi-stage GNN architecture that leverages hierarchical pooling and unpooling operations to progressively model global-to-local interactions across multiple spatial scales. We evaluate the proposed model on our newly developed CFD dataset simulating natural convection within a rectangular cavities with varying aspect ratios where the bottom wall is isothermal hot, the top wall is isothermal cold, and the two vertical walls are adiabatic. Experimental results demonstrate that the proposed model achieves higher predictive accuracy, improved training efficiency, and reduced long-term error accumulation compared to state-of-the-art (SOTA) GNN baselines. These findings underscore the potential of the proposed multi-stage GNN approach for modeling complex heat transfer in mesh-based fluid dynamics simulations. 

\end{abstract}

%% Keywords

\end{frontmatter}

\section{Introduction}
\label{sec:introduction}

Accurate heat transfer modeling and thermal analysis are foundational to modern science and engineering. Temperature fields govern material behavior, reaction rates, reliability, and energy efficiency. Their impact spans biomedicine~\cite{bounouar2016numerical,ragab2021heat,andreozzi2019modeling,basri2016computational}, electronics cooling ~\cite{murshed2017critical,nair2024comprehensive,yu2024comprehensive,arshad2024numerical,fazeli2024analysis}, automotive thermal management~\cite{falcone2021lithium,jiao2025cfd,bargal2025thermohydraulic,pasunurthi2024transient,bandhauer2011critical}, and manufacturing processes~\cite{norton2013computational,alteneiji2022heat,reynolds2023characterisation,hallam2004model,louhenkilpi1993real}, as well as core heat-transfer methodology and design practice~\cite{xia2018topology,deng2025topology}. Robust thermal models enable safer, more reliable, and more efficient systems across these domains. Computational Fluid Dynamics (CFD) provides numerical approaches for investigating fluid behavior and dynamics~\cite{versteeg2007introduction,pletcher2012computational}. Despite their effectiveness, conventional CFD methods rely heavily on precise physical modeling, detailed mesh generation, and substantial domain-specific expertise. In addition, these simulations are often computationally intensive, requiring significant processing power and memory resources.

Recent advances in deep learning (DL) have opened up new opportunities in the field of Computational Fluid Dynamics (CFD) by enabling data-driven models to learn complex nonlinear fluid behaviors directly from simulation or experimental data~\cite{vinuesa2022enhancing,kochkov2021machine}. Unlike traditional numerical solvers that rely on hand-crafted physical models and computationally expensive discretizations, DL-based approaches can approximate the underlying dynamics with significantly reduced computational cost, while capturing intricate spatial and temporal patterns, making them particularly attractive for real-time prediction, uncertainty quantification, and high-resolution flow analysis~\cite{sofos2025review,lye2020deep}.

DL approaches can be categorized into two main categrorirs including regular and irregular grid approaches. Lee et al.~\cite{lee2019data} represents an early effort to apply convolutional neural networks (CNNs) on regular grid structures for modeling transient fluid dynamics. A key innovation of this work is the incorporation of physics-based loss functions that explicitly enforce the conservation laws of mass and momentum. By embedding physical constraints directly into the learning process, this approach establishes a foundational link between machine learning and fluid mechanics, marking a major advancement in physics-informed modeling. Later, Wang et al. introduced Turbulent-Flow Net~\cite{wang2020towards}, which forecasts turbulent flow by learning its complex, nonlinear dynamics from spatiotemporal velocity fields generated by large-scale fluid simulations, with direct relevance to both turbulence and climate modeling. In another study, Ribeiro et al.~\cite{ribeiro2020deepcfd} proposed DeepCFD, a CNN-based framework for efficiently approximating solutions to non-uniform steady laminar flows. The model learns to predict full-field solutions of the Navier-Stokes equations, including both velocity and pressure components, directly from ground-truth data generated by a SOTA CFD solver. While these studies mainly focused on regular grid structures, real-world fluid dynamics are often based on irregular and non-uniform meshes that change with varying geometries and boundary conditions introducing significant challenges for model generalization, adaptability, and accurate representation of complex physical phenomena. To address this, more flexible approaches, such as Graph Neural Networks (GNNs)~\cite{scarselli2008graph,asif2021graph,pfaff2020learning}, have emerged, offering the ability to naturally handle irregular, and non-uniform geometries with varying mesh resolutions.

GNNs are a class of neural networks specifically designed to operate on mesh-structured data, where information is represented as nodes (vertices) connected by edges. In the context of a graph, each node represents an entity such as a point in space, a sensor, or a spatial location in a fluid and the edges encode the relationships or interactions between these entities. GNNs effectively learn from this irregular, non-Euclidean data structure by performing localized neural networks over the graph\cite{zhou2020graph,wu2020comprehensive}. At each layer of a GNN, a node updates its representation by aggregating features from its immediate neighbors, weighted by the graph’s adjacency structure. This neighborhood aggregation allows the network to gradually incorporate broader contextual information as layers deepen. As a result, GNNs can capture both local and global topological patterns inherent in the graph. This architecture makes GNNs particularly well-suited for tasks in CFD, where mesh-based simulations naturally form graphs with complex geometries and unstructured connectivity. By learning how physical quantities such as pressure, velocity, or temperature propagate across such topologies, GNNs improve the modeling of spatial-temporal dependencies and dynamic interactions within fluid systems\cite{wang2024recent}.

Among recent advances in GNN-based models, Graph Network-based Simulators (GNS), introduced by Sanchez-Gonzalez et al.\cite{sanchez2020learning}, and MeshGraphNets, introduced by Pfaff et al.\cite{pfaff2020learning}, have attracted considerable attention. These models represent physical systems as a graph, employing message passing and edge updating to simulate computational dynamics in mesh- and particle-based modeling. More specificly, GNS models the physical system as a graph, where particles are represented as nodes and their interactions are captured through learned message-passing mechanisms. MeshGraphNets presents a framework for learning mesh-based physical simulations, where the model is trained to perform message passing over mesh graphs while dynamically adapting the mesh discretization throughout the forward simulation process. Zhao et al.~\cite{zhao2023computationally} proposed a hybrid model combining convolutional and graph neural networks to overcome the limitations of GNNs in capturing complex topological structures, enabling a more effective representation of connectivity and flow pathways in porous media.

However, traditional GNNs still face limitations in effectively capturing  long-range interactions within complex graph structures. As message-passing mechanisms are typically restricted to a limited number of layers, the receptive field of each node often remains shallow, making it difficult for the network to access distant node information without degradation. On the other hand, increasing the number of layers to extend this receptive field can introduce challenges such as over-smoothing and vanishing gradients, where the gradients fail to propagate effectively during training. Our extensive experiments on high-resolution mesh structures in natural convection and vortex shedding simulations reveal significant challenges that underscore the limitations of existing methods in effectively capturing complex spatial dependencies~\cite{mendez2022challenges,wang2408recent}. These issues hinder GNNs’ ability to converge to optimal representations, especially in high-resolution CFD simulations where precise modeling of both localized flow variations and global structural influences is essential. Therefore, enhancing GNN architectures to better perform local detail extraction with global context integration remains a critical step toward applying GNN in fluid dynamics modeling.

Several recent studies have explored potential solutions to address this challenge. For instance, Lino et al.\cite{lino2022towards} proposed MultiScaleGNN, a novel multi-scale graph neural network designed to model unsteady continuum mechanics in scenarios involving multiple length scales and complex boundary conditions. Fortunato et al.\cite{fortunato2022multiscale} introduced a multi-scale GNN approach that relies on manually constructed coarser meshes to capture hierarchical features of the original geometry. Similarly, Li et al.~\cite{li2020multipole} proposed a multi-scale graph method by randomly pooling nodes and applying matrix factorization techniques to the adjacency matrix. Despite their contributions, these approaches exhibit notable limitations, such as the need for manual mesh generation, which introduces substantial computational overhead, or the use of random node selection in pooling operations, which increases the likelihood of generating artifacts and consequently degrades model performance.

In this work, we present a novel multi-stage Graph Neural Network (GNN) architecture designed to address the challenges of modeling heat transfer on high-resolution meshes. We introduce simple yet efficient pooling and un-pooling techniques mechanisms across multiple stages
which improve training efficiency, shortens training time, enhances predictive accuracy, and significantly mitigate error accumulation during long-term simulations. To construct a multi-stage architecture within our GNN model, we adopt the Pooling and Unpooling algorithms as introduced in the MAgNET~\cite{deshpande2022magnet} framework. The proposed architecture employs multiple parallel GNN branches, each processing a different resolution of the input mesh. Transitions between these resolutions are achieved using the pooling and un-pooling operators, which downsample and upsample the mesh structures, respectively. The multi-resolution features extracted from these branches are then fused and passed through a final high-resolution GNN refinement stage to predict the next time-step flow field. Note that the original implementations of Pooling \& Unpooling operations are computationally expensive and impose significant runtime overhead. To address this limitation, we develop an optimized implementation strategy that reduces computational cost while preserving the hierarchical representation power of the model.

To comprehensively assess the performance of the proposed approach, we constructed a new CFD dataset that captures the time-dependent evolution of temperature fields under natural convection within closed square cavities with varying aspect ratios where bottom wall and upper wall are isothermally hot and cold , respectively. We performed both quantitative and qualitative comparative analyses on this dataset. The experimental results demonstrate that the proposed model in this study not only achieves superior predictive accuracy but also significantly mitigates the accumulation of prediction errors over time during the testing phase. This translates to more stable and reliable long-term forecasting performance compared to standard Graph Neural Network (GNN) baselines, underscoring the model’s effectiveness in capturing intricate thermal-fluid interactions.

% \textcolor{red}{REGULAR OR IRREGULAR} \\
% We sum up the contributions of this paper as follows:
% \begin{itemize}
% \item We propose a multi-stage GNN architecture with pooling and unpooling operators for simulating buoyancy driven convection in rectangle cavities with varying aspect ratios., designed to effectively handle high-resolution meshes.
% \item We construct a new CFD dataset that simulates natural convection in a closed cavity with internal heat sources, capturing the time-dependent evolution of temperature fields.
% \end{itemize}

In the following, Section~\ref{Methodology} details the proposed model architecture. Section~\ref{Data preparation} describes the dataset and implementation details. Finally, Section~\ref{Results} presents the experiments and discusses the results.

        % \begin{subfigure}[t]{0.6\textwidth}
        % \centering
        %     \includegraphics[width=8cm, height=2cm]{ours_target_3_t.jpg}
        %     \caption{CFD Solver.}
        % \end{subfigure}

\begin{figure*}[t]
  \centering
  \begin{subfigure}[t]{1\textwidth}
    \centering
        \includegraphics[width=\textwidth]{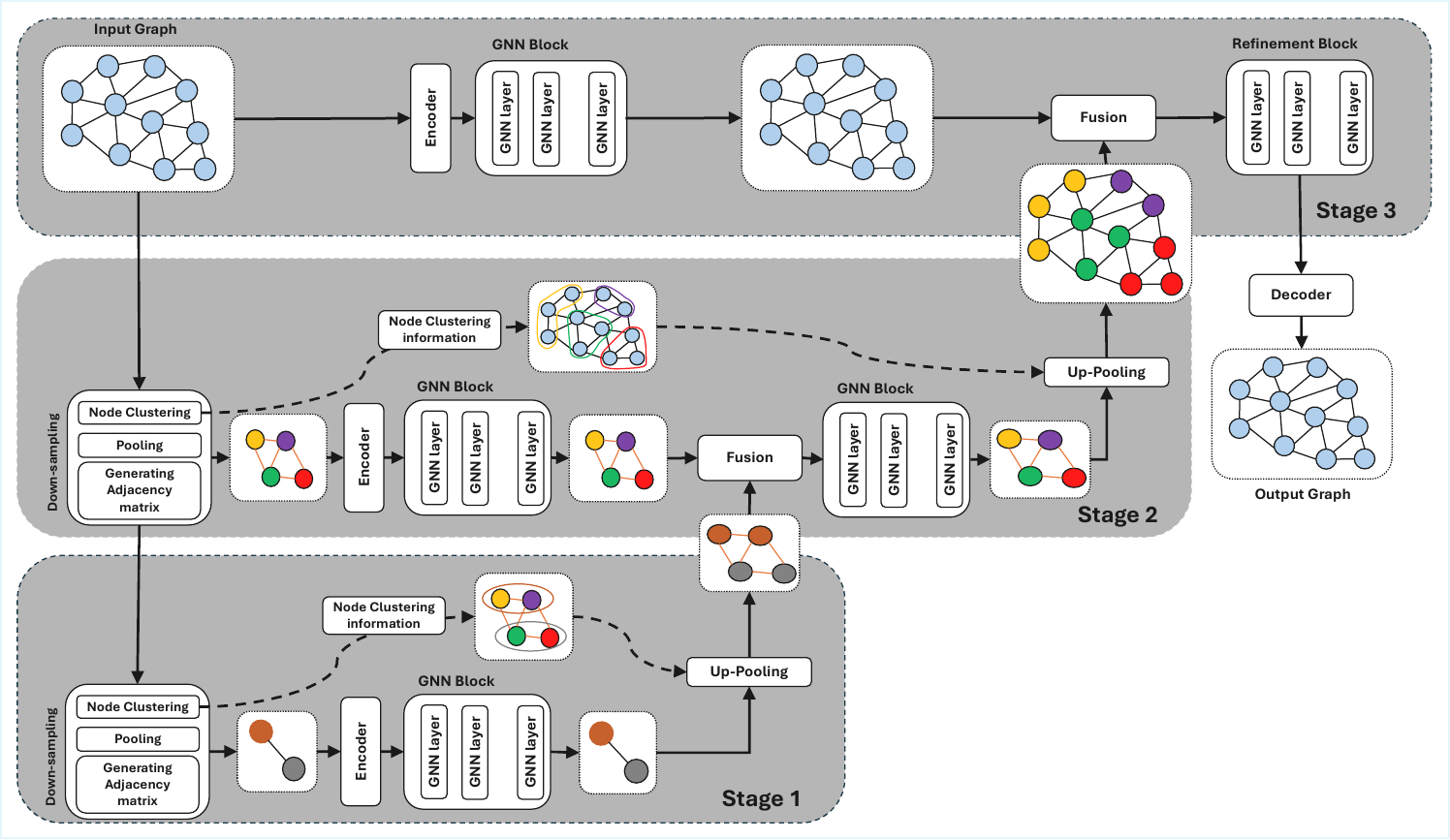}
        \caption{The overall structure of the proposed model, consisting of three stages. Nodes belonging to the same clique are indicated by the same color. The encoder is composed of multi-layer perceptrons (MLPs) that map the input space to a 128-dimensional feature space, while the decoder reconstructs the input space from the feature space. The aggregation module is also implemented using MLPs, concatenating input features of size 128 and transforming them into output features of the same size. The number of stages can vary depending on the size of the graph.}
  \end{subfigure}
  \label{fig:model}
\end{figure*}

\begin{figure*}[t]
  \centering
  \begin{minipage}[b]{1\textwidth}
    \centering
    \includegraphics[width=\textwidth]{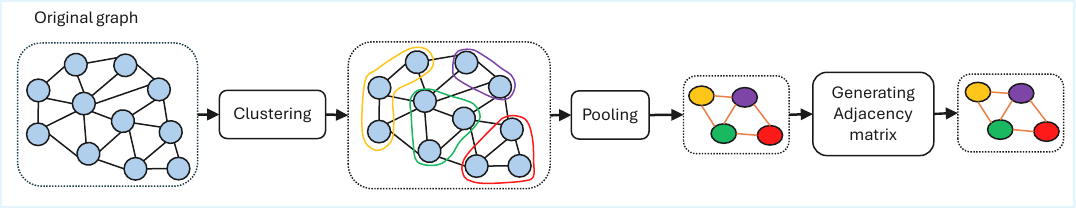}
    \centering
    \caption{The overall structure of the down-sampling algorithm begins with clustering the nodes in the original graph using a graph clustering algorithm~\ref{algorithm:1}. Subsequently, average pooling is applied within each clique to aggregate node features. Finally, a new adjacency matrix is constructed to define the connectivity of the resulting pooled graph.}
  \end{minipage}
\label{fig:pooling_}
\end{figure*}

\begin{figure*}[t]
  \centering
  \begin{minipage}[b]{1\textwidth}
    \centering
    \includegraphics[width=\textwidth]{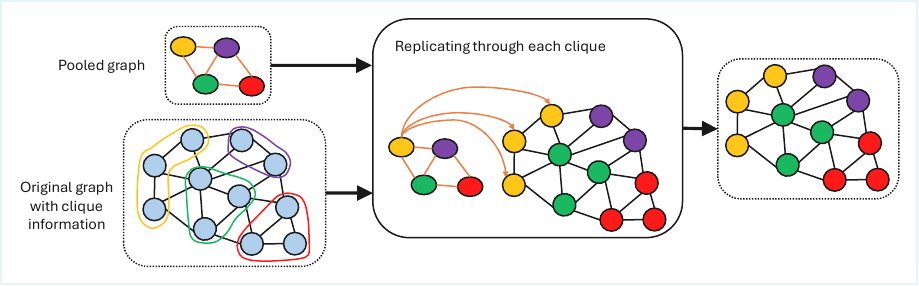}
    \centering
    \caption{Graph us-psampling (un-pooling) operation for reconstructing node-level features in the original high-resolution graph. Given the pooled graph and the clustering information, each node’s feature in the pooled graph is propagated to its corresponding clique in the original graph. This restores the original resolution while maintaining consistency with the pooled representation. The reconstructed node features are then combined, via concatenation or summation, with the original graph features to support hierarchical graph learning.}
  \end{minipage}
  \label{fig:unpooling}
\end{figure*}

\section{Methodology}
\label{Methodology}

This study introduces a multi-stage Graph Neural Network (GNN) framework for predicting heat transfer within a two-dimensional enclosed rectangle cavities where upper wall and bottom wall are cold and hot, respectively. The proposed model consists of three main components: GNN blocks, pooling and un-pooling operators, and a refinement block. The following sections provide a detailed description of each component.

\subsection{GNN block:}

Given a graph at time step $t$, denoted as $\mathcal{G}_t = (\mathcal{V}, \mathcal{E})$, each node $v_i \in \mathcal{V}$ has an input feature vector:

\begin{equation}
\label{input values}
\begin{aligned}
\mathbf{x}_i^t = \left[ \tau_i^t, \, \ell_i \right] \in \mathbb{R}^{d},
\end{aligned}
\end{equation}

where $\tau_i^t \in \mathbb{R}$ is the temperature of the $i^{th}$ node at time step t and and $\ell_i$ is the corresponding one-hot encoded label with length of 3, indicating the node type. Note that d = 4.  An encoder  maps raw input features into a latent space:

\begin{equation}
\label{initial encoder}
\begin{aligned}
\mathbf{h}_i^{(0)} = \phi_{\text{enc}}(\mathbf{x}_i^0),
\end{aligned}
\end{equation}

where $\mathbf{x}_i^0$ indicate raw input data and  $\phi_{\text{enc}}$ is a learnable transformation such as a multi-layer perceptron (MLP)~\cite{rumelhart1986learning}. The term $\mathbf{h}_i^{(0)}$ represents the latent features that serve as the input to the first layer of the GNN. The output dimension of $\mathbf{h}_i^{(0)}$ is set to 128 and remains fixed for all layers. For each GNN layer $l = 1, \dots, L$, node and edge features are updated using Eqs.~\ref{Edge update},~\ref{Message aggregation}, and~\ref{Node update} as follows:\\

\begin{equation}
\label{Edge update}
\begin{aligned}
\mathbf{e}_{ij}^{(l)} = \phi_e^{(l)} \left( \mathbf{h}_i^{(l-1)}, \mathbf{h}_j^{(l-1)},\mathbf{e}_{ij}^{(l-1)} \right),
\end{aligned}
\end{equation}

\begin{equation}
\label{Message aggregation}
\begin{aligned}
\mathbf{m}_i^{(l)} = \sum_{j \in \mathcal{N}(i)} \psi_m^{(l)} \left( \mathbf{e}_{ij}^{(l)} \right),
\end{aligned}
\end{equation}

\begin{equation}
\label{Node update}
\begin{aligned}
\mathbf{h}_i^{(l)} = \phi_n^{(l)} \left( \mathbf{h}_i^{(l-1)}, \mathbf{m}_i^{(l)} \right),
\end{aligned}
\end{equation}

\begin{figure*}[t]
    \centering
    \begin{subfigure}[b]{1\textwidth}
    \centering
    \includegraphics[width=\textwidth]{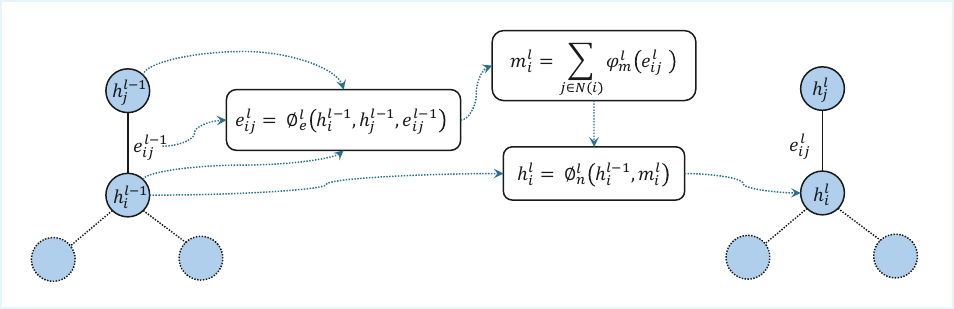}
    \centering
    \end{subfigure}
    \caption{Illustration of the node and edge update process across hierarchical layers in a graph.}
\label{fig:updating}
\end{figure*}

where $\phi_e^{(l)}, \phi_n^{(l)}, \psi_m^{(l)}$ are learnable functions (e.g., MLPs). The term $\mathbf{m}_i^{(l)}$ indicates the aggregated message vector to node $i$ at GNN layer $l$. This vector summarizes the influence of node $i$’s neighbors on its next state. $ \mathcal{N}(i)$ indicates a  set of neighboring nodes connected to node $i$. The neighborhood set $ \mathcal{N}(i)$ is derived from the sparse adjacency matrix provided by the mesh data . The term $\mathbf{e}_{ij}^{(l)}$ also shows the edge feature from node $j$ to node $i$, computed in the edge update step.  The edge feature $\mathbf{e}_{ij}^{(l)}$ represents the information propagated from node $j$ to node $i$ at layer $l$, and is computed during the edge update step.  $\psi^{(l)}$ denotes the message transformation function at layer $l$, typically implemented as a learnable multilayer perceptron (MLP) or a linear transformation, and applied to the edge features to encode the interaction information between connected nodes. $\mathbf{h}_i^{(l)}$ represents the updated feature vector of node $i$ at layer $l$ of the GNN, capturing its new state after aggregating information from neighboring nodes. The update is governed by the node update function $\phi_h^{(l)}$, typically implemented as a learnable multilayer perceptron (MLP), which takes as input the node’s previous hidden state $\mathbf{h}_i^{(l-1)}$ along with the aggregated messages. Fig.~\ref{fig:updating} illustrates the process of node and edge feature updates. The decoder transforms the fused latent representation \(\mathbf{h}_i^{L}\) into the predicted physical quantities at the next time step:

\begin{equation}
\begin{aligned}
\hat{\tau}_i^{t+1} = \phi_{\mathrm{dec}} \left( \mathbf{h}_i^{L} \right),
\end{aligned}
\end{equation}

where $\tau_i^{t+1}$ the temperature of the $i^{th}$ node at time step t+1, \(\phi_{\mathrm{dec}}\) is a trainable multilayer perceptron (MLP).

In the following, we present a multi-stage architecture that spatially subsamples the original graph to capture global interactions between distant nodes. This is achieved through the introduction of pooling and unpooling modules, which enable hierarchical processing within the model.

\subsection{Pooling Layer :}

To reduce the graph resolution while preserving structural connectivity, we employ a pooling strategy that groups nodes into cliques. Given an input graph with \( N \) nodes and sparse adjacency matrix \( \mathbf{A} \in \{0,1\}^{N \times N} \), we partition the set of nodes into \( \tilde{N} \) non-overlapping subsets \( \mathcal{S}_1, \mathcal{S}_2, \ldots, \mathcal{S}_{\tilde{N}} \), where each \( \mathcal{S}_i \) forms a fully-connected subgraph (i.e., a clique), and \( |\mathcal{S}_i| \in \{2, 3\} \).  Algorithm~\ref{algorithm:1} outlines the procedure for selecting cliques from the original graph. Once the cliques belonging to the pooled graph are identified, the corresponding pooled adjacency matrix must be constructed to represent the interconnections among these cliques.  Algorithm~\ref{algorithm:2} describes the construction of the pooled graph's adjacency matrix. Specifically, it ensures that any pair of cliques in the pooled graph are connected in the new adjacency matrix $\tilde{\mathbf{A}}$ if there exists at least one edge between their constituent nodes in the original adjacency matrix $\mathbf{A}$.  The resulting pooled adjacency matrix \( \tilde{\mathbf{A}} \in \{0,1\}^{\tilde{N} \times \tilde{N}} \) is computed as:
\[
\tilde{A}_{ij} = \begin{cases}
1, & \text{if } \exists \, u \in \mathcal{S}_i, \, v \in \mathcal{S}_j \text{ such that } A_{uv} = 1 \\
0, & \text{otherwise}
\end{cases}
\]

This method preserves inter-group connectivity while reducing the graph's complexity, and can be implemented in a stochastic but reproducible manner using a random seed. Fig.~\ref{fig:pooling_} illustrates the clustering, pooling, and generating adjacency matrix.

%%%%%%%%%%%%%%%%%%%%%%%%%%%%%%%%%%%%%%%%%%%%%%%%%%%%%%%%%

\begin{algorithm}
\caption{Subgraph Pooling from a Parent Graph}
\begin{algorithmic}[1]
\Require Adjacency matrix $\mathbf{A} \in \mathbb{R}^{N \times N}$
\Ensure Set of subgraphs $\mathcal{S}$ and pooled adjacency matrix $\tilde{\mathbf{A}} \in \mathbb{R}^{\tilde{N} \times \tilde{N}}$

\State $\mathcal{S} \gets \emptyset$ \Comment{Initialize subgraph container}
\State $\mathcal{V} \gets \{1, 2, \dots, N\}$ \Comment{Set of available nodes}
\State $\mathcal{G}$ \Comment{temporary subgraph}
\State $\mathbf{A}_{\text{temp}} \gets \mathbf{A}$ \Comment{Create a mutable copy of $\mathbf{A}$}

\While{$\mathcal{V} \neq \emptyset$}
    \State Randomly choose a node $v$ from $\mathcal{V}$
    \State $\mathcal{G} \gets \{v\}$ \Comment{Start new subgraph with node $v$}
    \State $\mathcal{N}_v \gets \{u \in \mathcal{V} \mid u \neq v \ \text{and} \ \mathbf{A}_{\text{temp}}[u, v] = 1\}$

    \For{each node $u$ in $\mathcal{N}_v$}
        \If{$\forall w \in \mathcal{G}, \ \mathbf{A}_{\text{temp}}[w, u] = 1$}
            \State $\mathcal{G} \gets \mathcal{G} \cup \{u\}$
        \EndIf
    \EndFor

    \State $\mathcal{V} \gets \mathcal{V} \setminus \mathcal{G}$ \Comment{Exclude processed nodes}

    \For{each $w \in \mathcal{G}$}
        \State Set $\mathbf{A}_{\text{temp}}[w, :] \gets 0$ and $\mathbf{A}_{\text{temp}}[:, w] \gets 0$
    \EndFor

    \State Append $\mathcal{G}$ to $\mathcal{S}$: $\mathcal{S} \gets \mathcal{S} \cup \{\mathcal{G}\}$
\EndWhile

\State Compute pooled matrix $\tilde{\mathbf{A}}$ from $\mathcal{S}$ 

\end{algorithmic}
\label{algorithm:1}
\end{algorithm}

%%%%%%%%%%%%%%%%%%%%%%%%%%%%%%%%%%%%%%%%%%%%%%%%%%5

\begin{algorithm}
\caption{Construct Pooled Adjacency Matrix from Subgraphs}
\begin{algorithmic}[1]
\Require Original adjacency matrix $\mathbf{A} \in \mathbb{R}^{N \times N}$; list of subgraphs $\mathcal{S} = \{\mathcal{S}_1, \dots, \mathcal{S}_{\tilde{N}}\}$
\Ensure Pooled adjacency matrix $\tilde{\mathbf{A}} \in \mathbb{R}^{\tilde{N} \times \tilde{N}}$

\State $\tilde{N} \gets |\mathcal{S}|$ \Comment{Number of pooled nodes}
\State $\tilde{\mathbf{A}} \gets \mathbf{0}^{\tilde{N} \times \tilde{N}}$ \Comment{Initialize zero matrix}

\For{$r = 1$ to $\tilde{N}$}
    \For{$c = 1$ to $\tilde{N}$}
        \If{there exists $n \in \mathcal{S}_r$ and $m \in \mathcal{S}_c$ such that $\mathbf{A}[n, m] = 1$}
            \State $\tilde{\mathbf{A}}[r, c] \gets 1$
        \EndIf
    \EndFor
\EndFor

\end{algorithmic}
\label{algorithm:2}
\end{algorithm}

%%%%%%%%%%%%%%%%%%%%%%%%%%%%%%%%%%%%%%%%%%%%%%%%%%

%%   Nodes ,... Super nodes 
\subsection{Un-pooling layer:}
To reconstruct node-level features in the upper (original) graph from the pooled graph, we introduce a graph unpooling operation inspired by the unpooling mechanism in~~\cite{deshpande2022magnet}. In this process, the feature of each node in the pooled graph is uniformly propagated to all nodes within its corresponding subgraph (clique) in the higher-resolution graph. That is, each node in a subgraph inherits the same feature representation from its associated pooled node. This operation restores the original graph resolution while maintaining consistency with the pooled representation, thereby enabling effective feature reconstruction and facilitating hierarchical graph learning. The nodes of unpooled graph are concatenated or summed with nodes in original graph. 

%%%%%%%%%%%%%%%%%%%%%%%%%%%%%%%%%%%%%%%%%%%%%%%%%%

\subsection{Fast pooling and un-pooling implementation}
The original pooling and unpooling implementations are computationally expensive, which slows down training and reduces overall efficiency~\cite{deshpande2022magnet}. To address this, we design optimized pooling and unpooling operators that leverage fixed-size subgraph indexing and tensor-based aggregation, avoiding costly for-loops. This approach significantly reduces computational overhead while preserving flexibility in handling variable-sized clusters. Detailed descriptions and implementations are provided in~\ref{appendix:pooling_unpoolin}.

\subsection{Data preparation:}
\label{Data preparation}

In our dataset, we consider buoyancy-driven natural convection in sealed rectangular cavities with varying aspect ratios, from 1:1 (square) to 1:4 (horizontally elongated), discretized on structured quadrilateral meshes. The thermal boundary conditions are fixed: the bottom wall is isothermal hot, the top wall is isothermal cold, and the vertical sidewalls are adiabatic. Varying the aspect ratio profoundly alters the flow organization and heat-transfer pathways, yielding different Rayleigh-Bénard cell counts, spacings, and intensities, and thus a wider range of temperature gradients to learn.
Baseline GNNs performed adequately when trained and tested on a single aspect ratio, but their learning destabilized and test accuracy dropped sharply when trained across multiple aspect ratios, revealing a sensitivity to geometric variability. In contrast, the proposed multi-stage model in this study maintains robust generalization across aspect ratios, indicating it captures the underlying buoyant transport mechanisms rather than memorizing geometry-specific patterns.
To more closely reflect practical simulations, we emphasize higher-resolution meshes than those used in prior public datasets, enabling finer near-wall thermal layers and sharper cell boundaries. Motivated by these challenges, we construct a new dataset coupling high-resolution meshes with a diverse set of aspect ratios; to the best of our knowledge, this is the first such dataset used to train and evaluate graph neural networks for natural-convection cavities with hot-bottom/cold-top and adiabatic sidewalls.
The dataset is produced by solving the incompressible Navier-Stokes equations for buoyancy-driven flow under the Boussinesq approximation, coupled with the energy equation for temperature. Fluid density is treated as constant except in the buoyancy term, where it varies linearly with temperature:

\begin{equation}
\begin{aligned}
\rho(T) = \rho_{0} \left[ 1 - \beta \left( T - T_{0} \right) \right],
\end{aligned}
\end{equation}

The non-dimensional groups are the Prandtl number $Pr = \frac{\nu}{\alpha}$, and Rayleigh number $Ra = \frac{g \, \beta \, \Delta T \, H^{3}}{\nu \, \alpha}$ with $\Delta T = T_{\text{hot}} - T_{\text{cold}}$. In this work $P_r=1$.

The governing equations consist of continuity, momentum and energy equations are as follows:

\begin{align}
\nabla \cdot \mathbf{u} &= 0, \\
\rho \left( \frac{\partial \mathbf{u}}{\partial t} + \mathbf{u} \cdot \nabla \mathbf{u} \right) &= -\frac{1}{\rho_0} \nabla p + \nu \nabla^2 \mathbf{u} + \beta (T - T_0)\mathbf{g}, \\
\frac{\partial T}{\partial t} + \mathbf{u} \cdot \nabla T &= \alpha \nabla^2 T.
\end{align}

where $\rho_0$ denotes the reference density, and $\nu = \mu / \rho_0$ is the kinematic viscosity. Furthermore, $\alpha$, $\beta$, and $\mathbf{g}$ represent the thermal diffusivity, thermal expansion, and gravitational acceleration, respectively. The CFD simulations were performed using the cell-centered Finite Volume Method (FVM) with a segregated PIMPLE algorithm, as implemented in OpenFOAM v12.

A constant-property Boussinesq fluid was used strictly for data generation, not intended to represent a specific material. In the simulation, $\rho_0$, $\nu$, $\alpha$, $\beta$, and $T_0$ are $1000 \, \text{kg/m}^3$, $0.001 \, \text{m}^2/\text{s}$, $0.001 \, \text{m}^2/\text{s}$, $0.001 \, \text{K}^{-1}$, and $300 \, \text{K}$, respectively. Furthermore, gravity is imposed in a downward vertical direction ($g = 9.81$). CFD simulations were conducted for 2D rectangular cavities with different aspect ratios ($L/H$). The height of the cavity was kept constant ($H = 1 \, \text{m}$) and the widths ($L$) were changed.

\begin{equation}
\begin{aligned}
L/H = AR \in \{1,2,3,4\}.
\end{aligned}
\end{equation}

An isothermal condition ($T = 300 \, \text{K}$) was assumed for the top wall. Both side walls were assumed to be adiabatic. The bottom wall was assumed to be an isothermal hot boundary and its temperature in different simulations varied between $300.8$ to $302 \, \text{K}$. A no-slip boundary condition was imposed for all walls. A uniform grid size ($H/40$) was employed for all simulations. Hence, the employed grids had $1600$, $3200$, $5400$, and $6400$ cells when changing the aspect ratio from $1$ to $4$, respectively.

\section{Experiment and Results:}
\label{Results}

%%%%%%%%%%%%%%%%%%%%%%%%%%%%%%%%%%%
%%%%%%%%%%%%%%%%%%%%%%%%%%%%%%%%%%%

\begin{figure*}[t]
    \centering

        \begin{subfigure}[t]{0.6\textwidth}
        \centering
            \includegraphics[width=4cm, height=2cm]{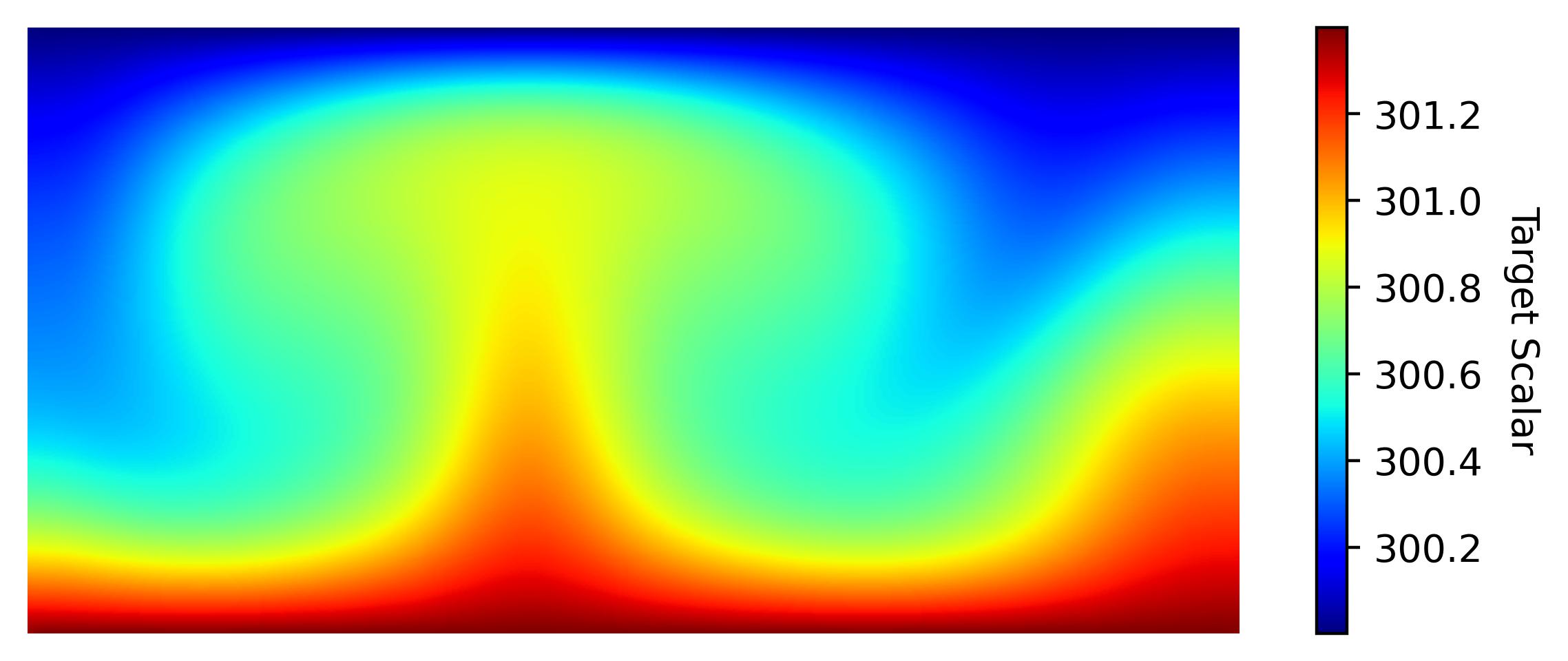}
            \caption{CFD Solver.}
        \end{subfigure}
        \centering
        \begin{subfigure}[t]{0.6\textwidth}
        \centering
            \includegraphics[width=4cm, height=2cm]{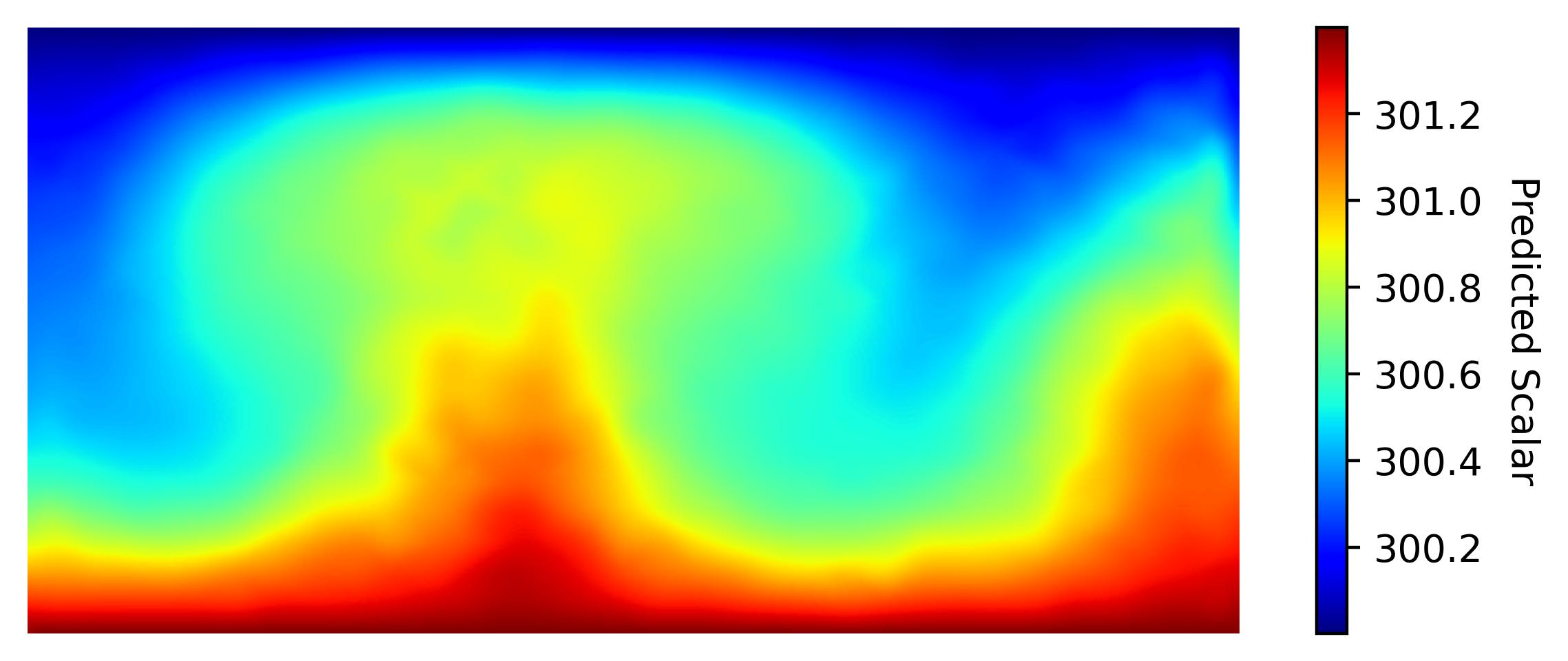}
            \caption{MeshGraphNets.}
        \end{subfigure}
        \centering
        \begin{subfigure}[t]{0.6\textwidth}
        \centering
            \includegraphics[width=4cm, height=2cm]{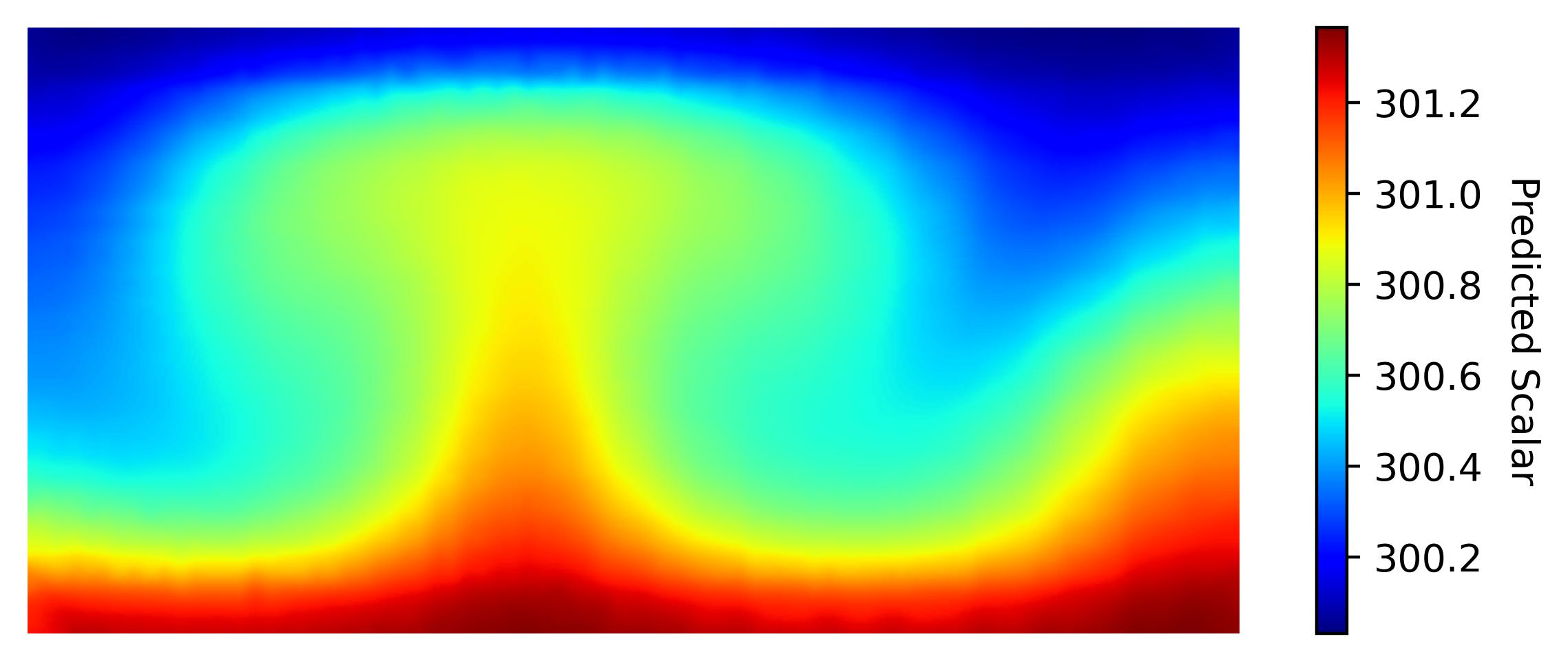}
            \caption{Proposed Model.}
        \end{subfigure}

    \caption{Contours of temperature as predicted by (a) CFD solver, (b) MeshGraphNets, and (c ) the proposed model for aspect ratio of 2, T-hot =301.4K and T-cold=300K.}
\label{fig:results_temp_as_12}
\end{figure*}

\begin{figure*}[t]
    \centering

        \begin{subfigure}[t]{0.6\textwidth}
        \centering
            \includegraphics[width=8cm, height=2cm]{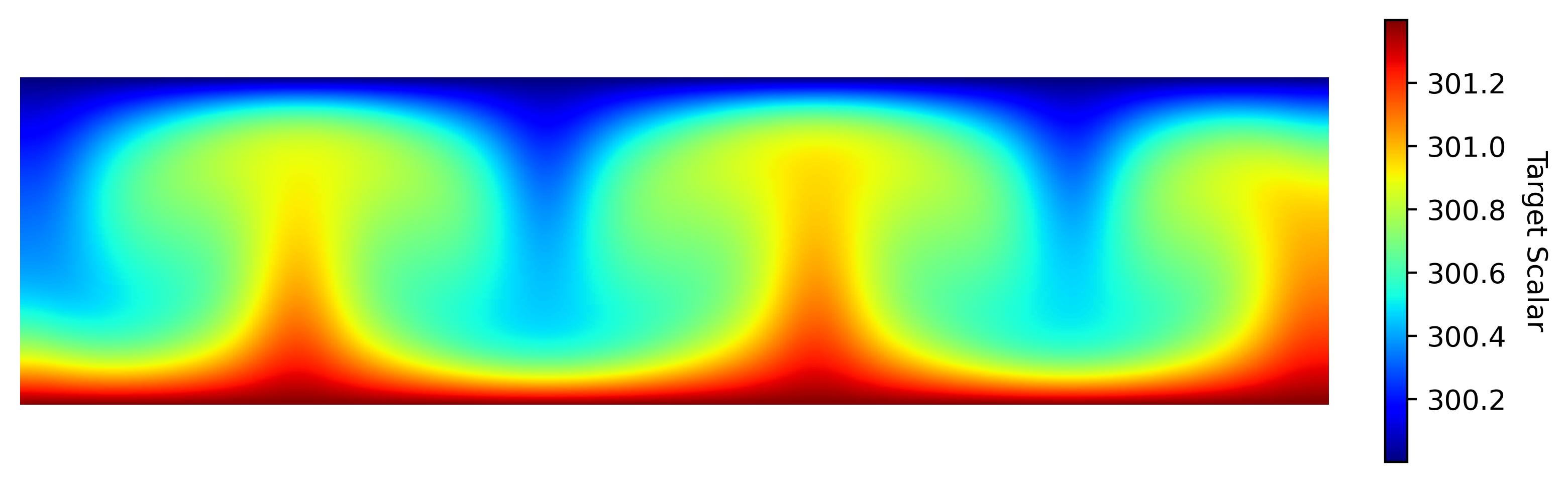}
            \caption{CFD Solver.}
        \end{subfigure}
        \centering
        \begin{subfigure}[t]{0.6\textwidth}
        \centering
            \includegraphics[width=8cm, height=2cm]{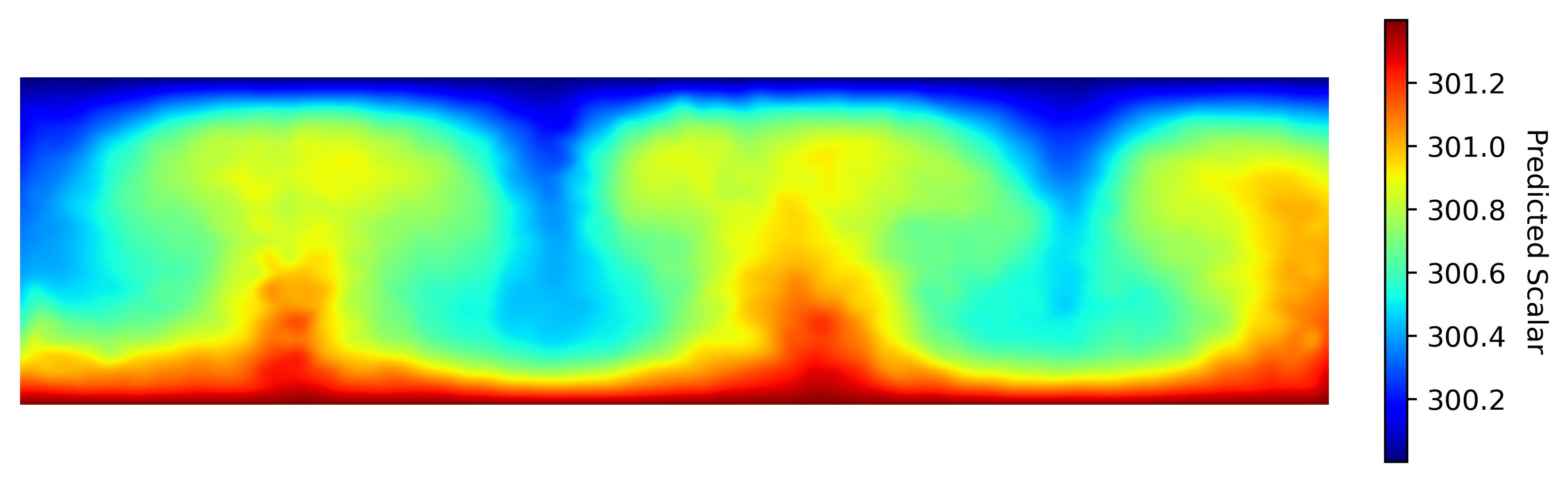}
            \caption{MeshGraphNets.}
        \end{subfigure}
        \centering
        \begin{subfigure}[t]{0.6\textwidth}
        \centering
            \includegraphics[width=8cm, height=2cm]{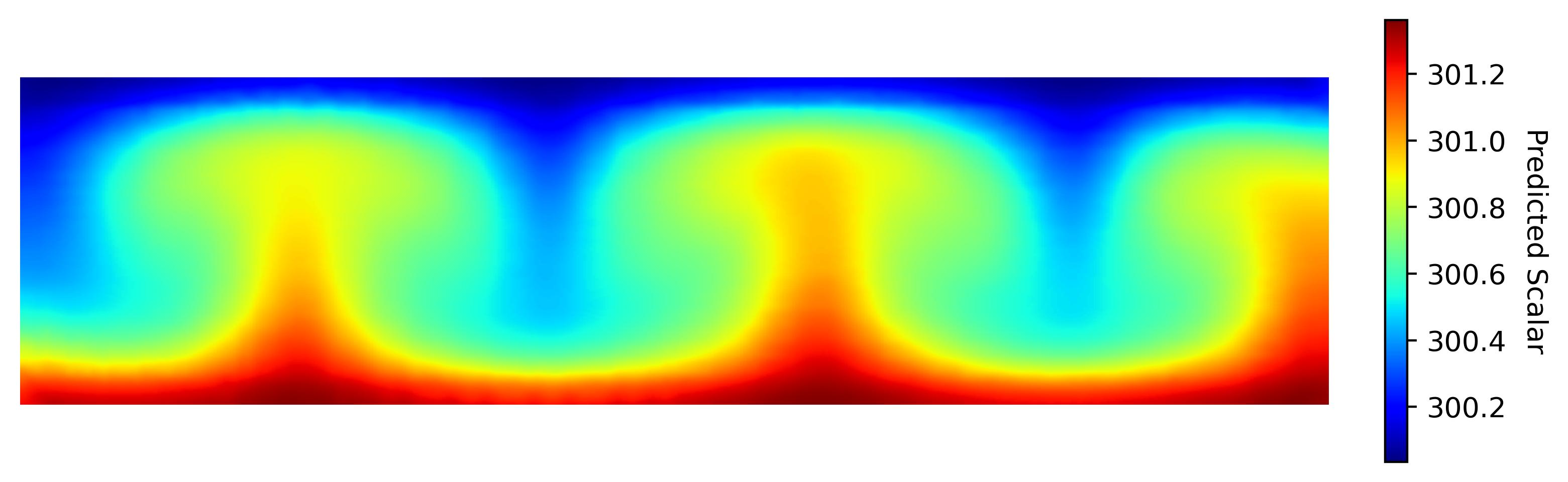}
            \caption{Proposed Model.}
        \end{subfigure}

    \caption{Contours of temperature as predicted by (a) CFD solver, (b) MeshGraphNets, and (c ) the proposed model for aspect ratio of 4, T-hot =301.4K and T-cold=300K.}
\label{fig:results_temp_as_14}
\end{figure*}

\begin{figure*}[t]
    \centering
    \begin{subfigure}[t]{1\textwidth}
        \centering
        \begin{subfigure}[t]{0.45\textwidth}
            \centering
            \includegraphics[width=3cm, height=2cm]{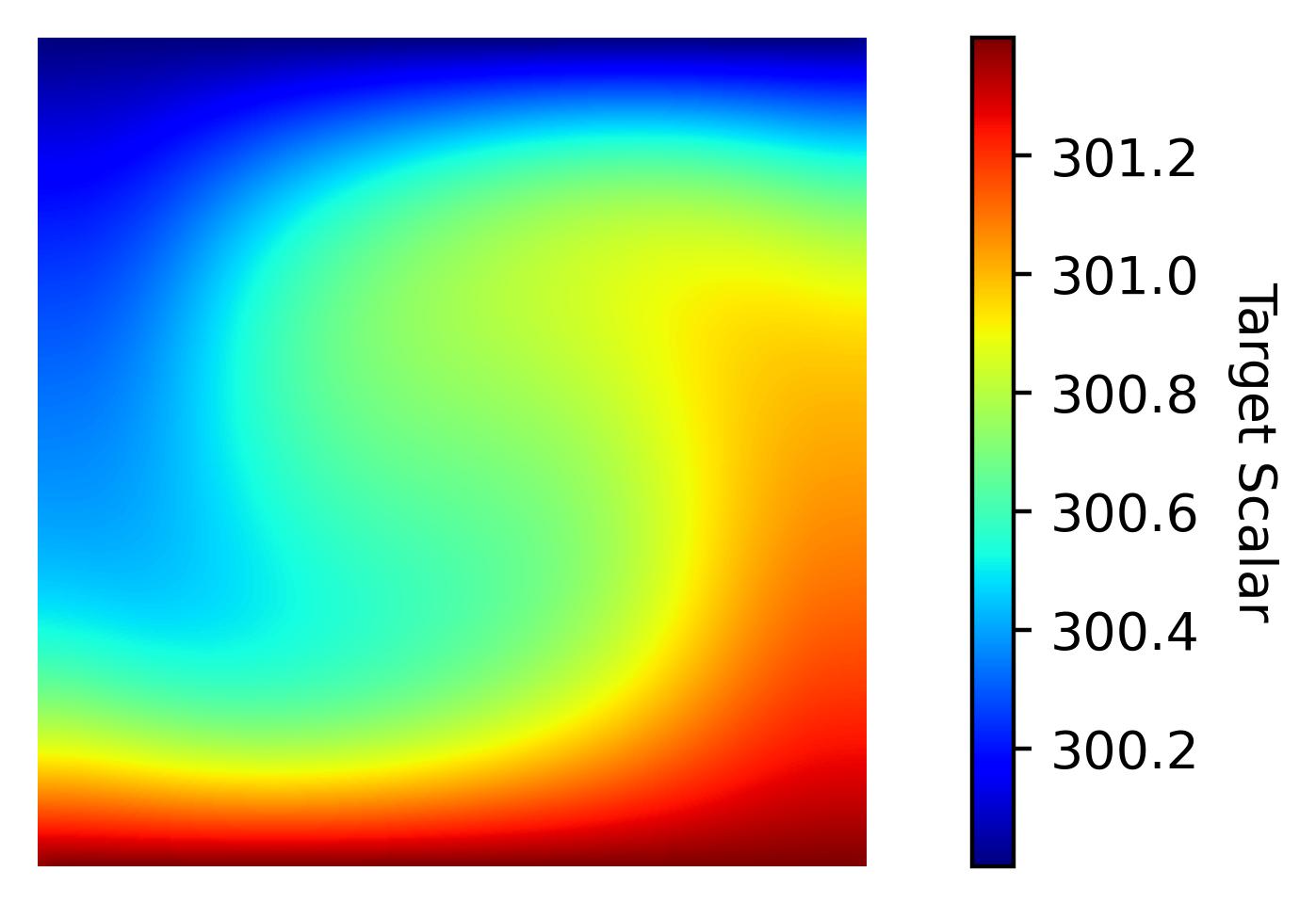}
            \caption{CFD Solver.}
        \end{subfigure}
        \centering
        \begin{subfigure}[t]{0.45\textwidth}
            \centering
            \includegraphics[width=3cm, height=2cm]{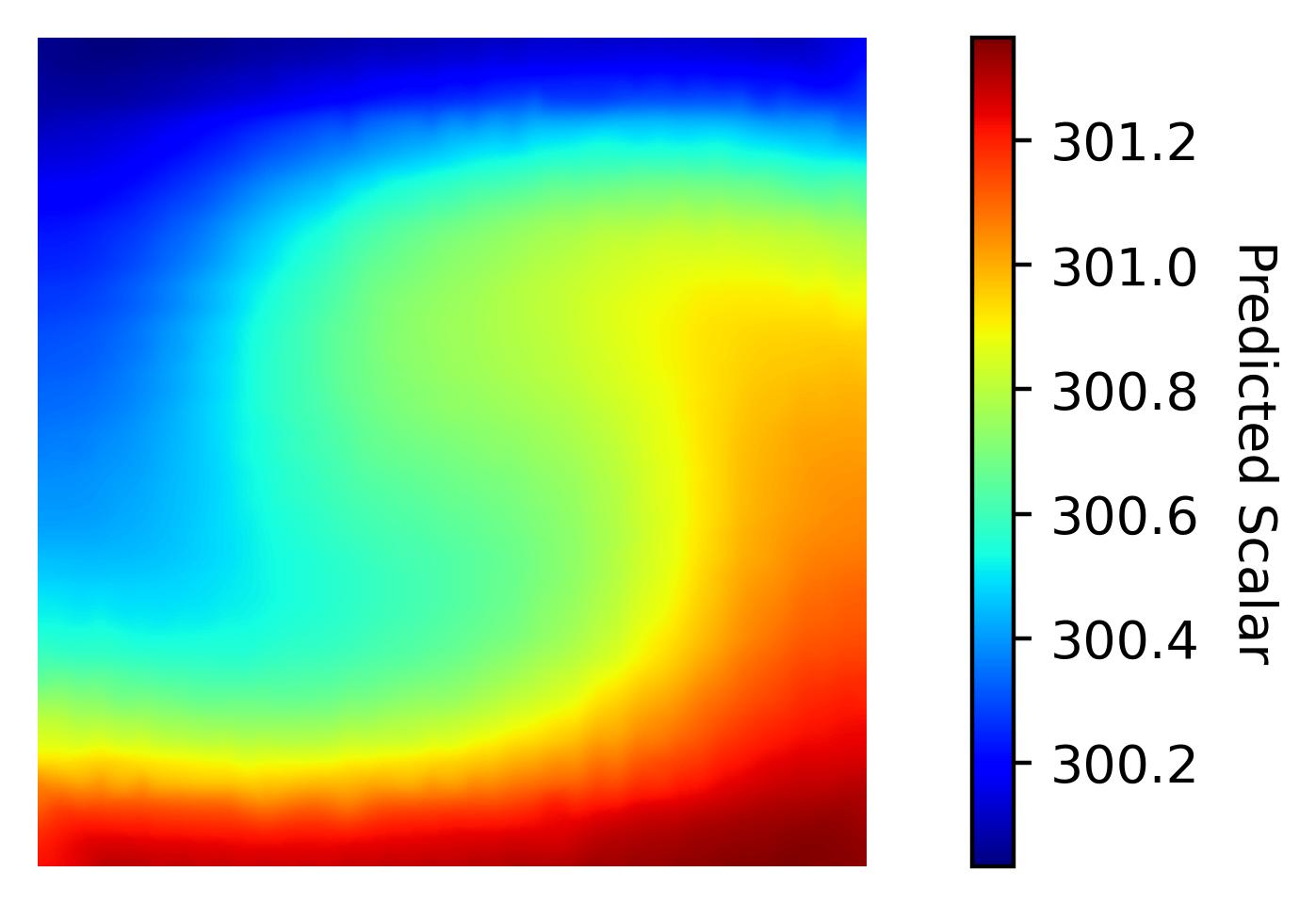}
            \caption{Proposed Model.}
        \end{subfigure}
        \centering
    \end{subfigure}
    \caption{Contours of temperature as predicted by CFD solver (a)  and the proposed model  (b ) for aspect ratio of 1, T-hot =301.4K and T-cold=300K.}
\label{fig:results_temp_as_11}
\end{figure*}

\begin{figure*}[t]
    \centering
    \begin{subfigure}[t]{1\textwidth}
        \centering
        \begin{subfigure}[t]{0.5\textwidth}
            \centering
            \includegraphics[width=6cm, height=2cm]{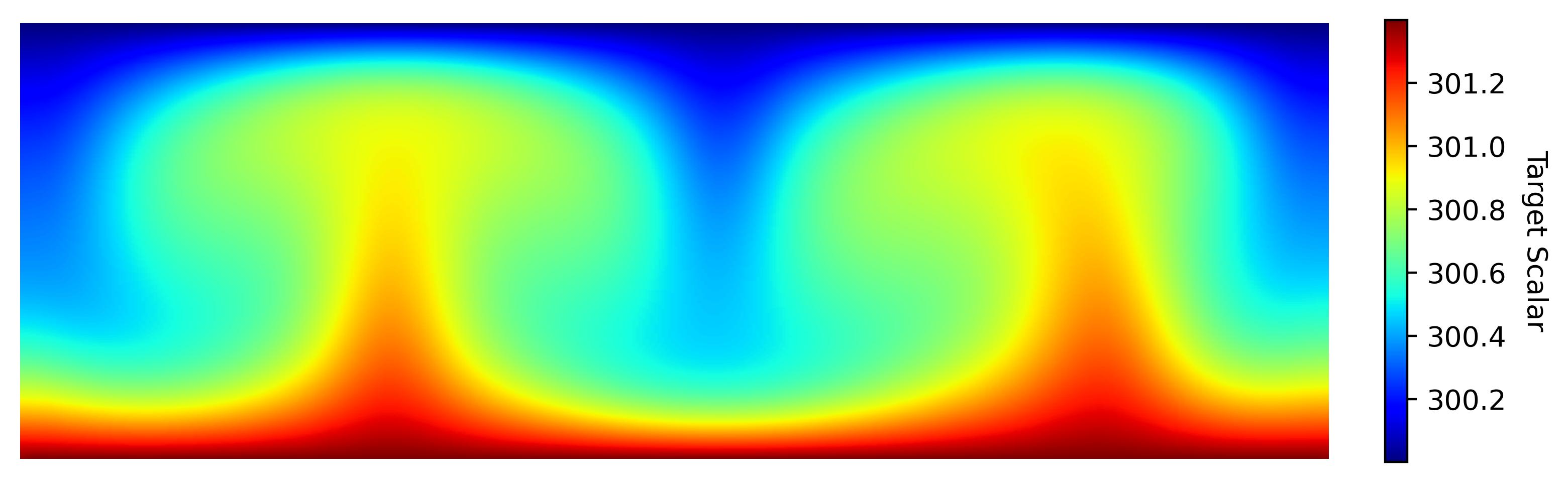}
            \caption{CFD Solver.}
        \end{subfigure}
        \centering
        \begin{subfigure}[t]{0.5\textwidth}
            \centering
            \includegraphics[width=6cm, height=2cm]{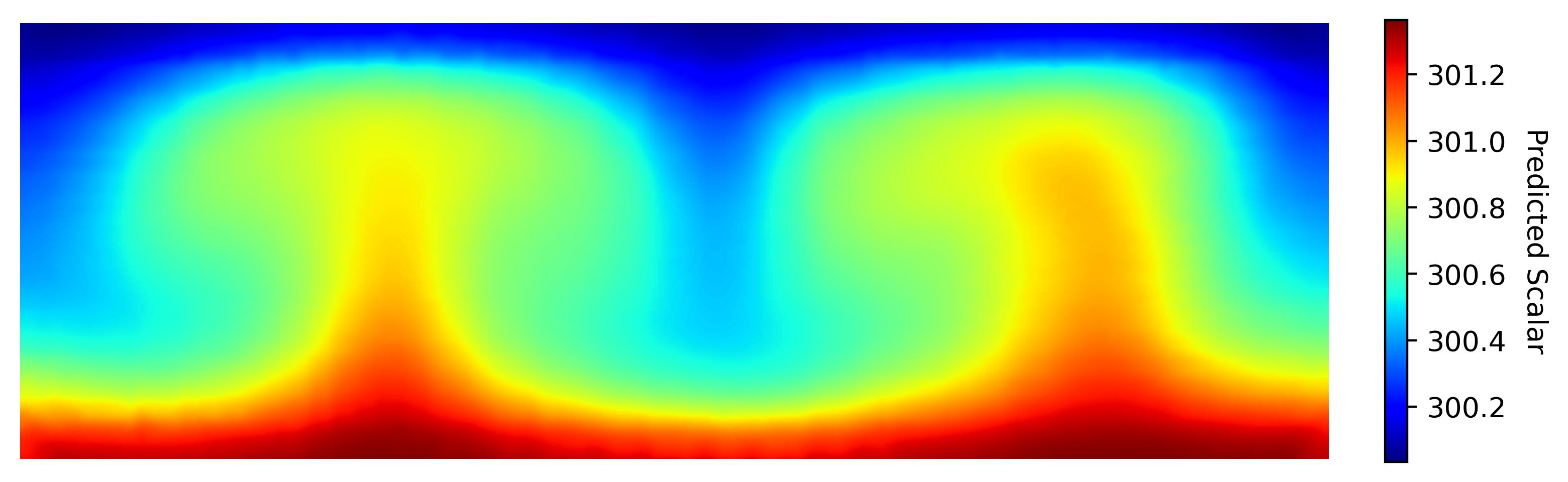}
            \caption{Proposed Model.}
        \end{subfigure}
    \end{subfigure}
    \caption{Contours of temperature as predicted by (a) CFD solver, and (b ) the proposed model for aspect ratio of 3, T-hot =301.4K and T-cold=300K.}
\label{fig:results_temp_as_13}
\end{figure*}

%%%%%%%%%%%%%%%%%%%%%%%%%%%%%%%%%%%
%%%%%%%%%%%%%%%%%%%%%%%%%%%%%%%%%%%

In this section, we conduct qualitative and quantitative analyses on the natural convection dataset, comparing the performance of conventional graph neural networks (GNNs) with our proposed model. Specifically, we benchmark our multi-stage deep GNN against MeshGraphNets~\cite{pfaff2020learning, li2020accelerating}, a GNN-based baseline widely used for mesh-based physical simulations.

As an accuracy metric, we use the Mean Squared Error (MSE) between the predicted mesh outputs and the corresponding ground truth solutions generated by the CFD solver (OpenFOAM). In addition, we employ SSIM as the second metric to assess the structural consistency of the predicted flow fields. SSIM provides insight into how well the models capture and preserve the underlying flow patterns during prediction.

It is important to highlight that MeshGraphNets was trained on a lower-resolution mesh compared to our proposed model. As discussed in the introduction, the architecture of MeshGraphNets faces challenges when handling high-resolution mesh data, primarily due to limited scalability and inefficiencies in its message passing mechanism. For example, the mesh for the 1:4 aspect ratio enclosure in our dataset contains approximately 6,650 nodes. To make training feasible with MeshGraphNets, we down-sampled the mesh to approximately 1,750 nodes by applying a minimum distance threshold between neighboring nodes. Therefore, in the experimental evaluation, we report the results for MeshGraphNets based on this reduced-resolution setting. Moreover, we attempted to train MeshGraphNets on the entire dataset containing enclosures with varying aspect ratios. However, MeshGraphNets did not converge to a stable solution under this setting. As a result, we limited MeshGraphNets training to only two aspect ratios. In contrast, our proposed model demonstrated the capability to train on the full dataset, covering all aspect ratios at full mesh resolution.

Figs.\ref{fig:results_temp_as_12} and \ref{fig:results_temp_as_14} present the predicted results alongside the corresponding CFD solver solutions for enclosures with aspect ratios of 1:2 and 1:4 under identical temperature boundary conditions. For instance, in Fig.\ref{fig:results_temp_as_12} compares the temperature contours as predicted by CFD solver, MeshGraphNets, and the proposed model in this study, respectively. The obtained results clearly indicate that the proposed model in this study outperforms MeshGraphNets, capturing fine-grained spatial features and closely matching the CFD solver predictions. Due to training stability issues, MeshGraphNets was trained on a reduced-resolution version of the dataset. Additional results for aspect ratios of 1:1 and 1:3 are shown in Fig.~\ref{fig:results_temp_as_11} and Fig.~\ref{fig:results_temp_as_13}, respectively.

\begin{figure*}[h]

        \centering
        \begin{subfigure}[t]{0.6\textwidth}
        \centering
            \includegraphics[width=8cm, height=2cm]{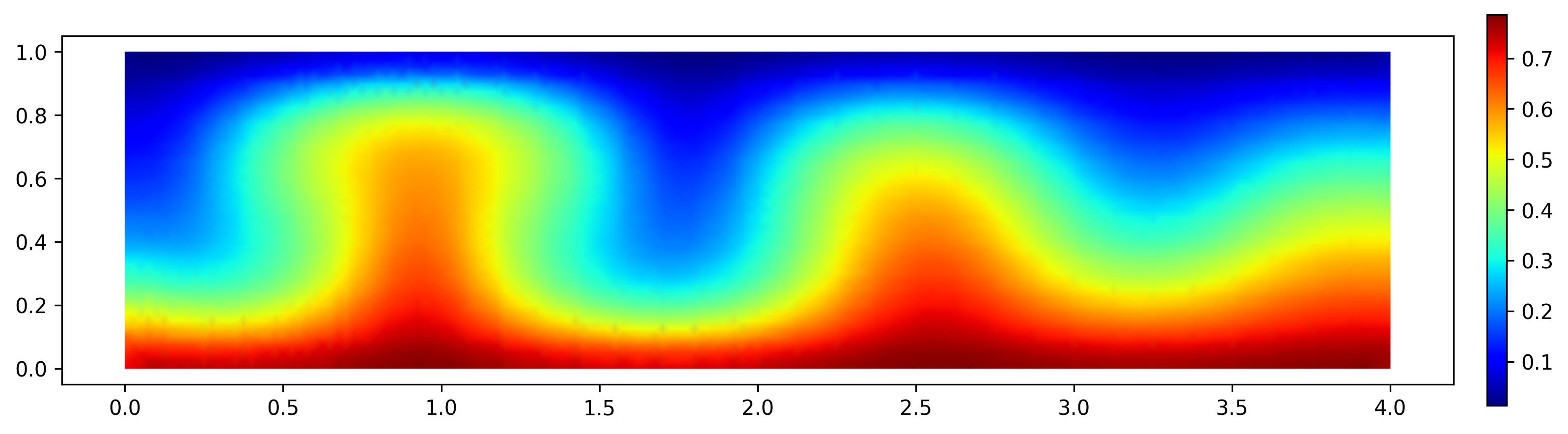}
            \caption{CFD Solver}
        \end{subfigure}
        \begin{subfigure}[t]{0.6\textwidth}
        \centering
            \includegraphics[width=8cm, height=2cm]{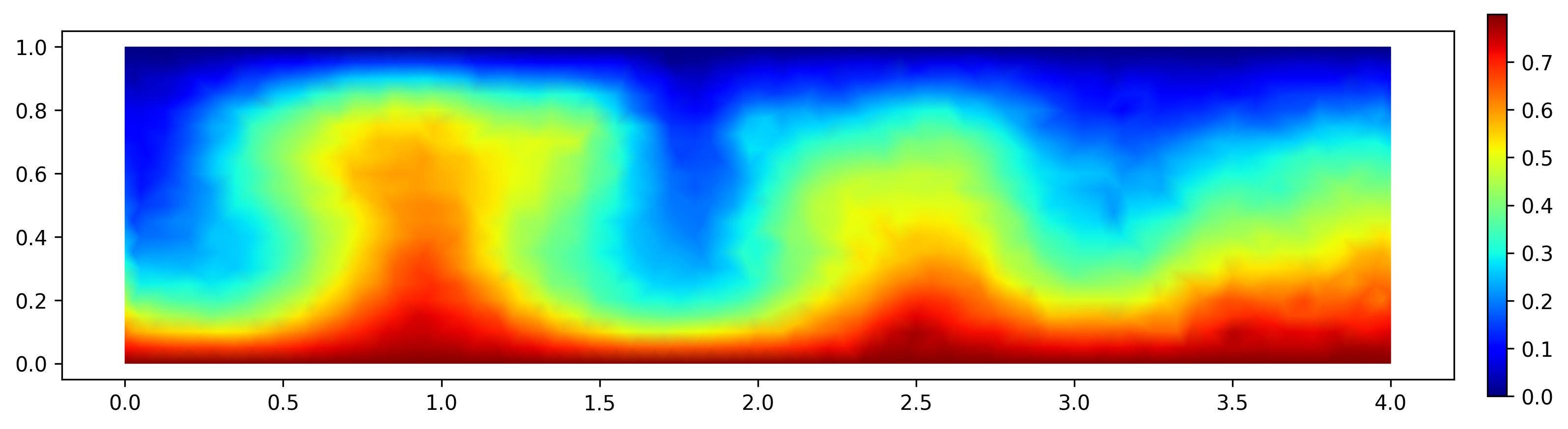}
            \caption{MeshGraphNets}
        \end{subfigure}
        \begin{subfigure}[t]{0.6\textwidth}
        \centering
            \includegraphics[width=8cm, height=2cm]{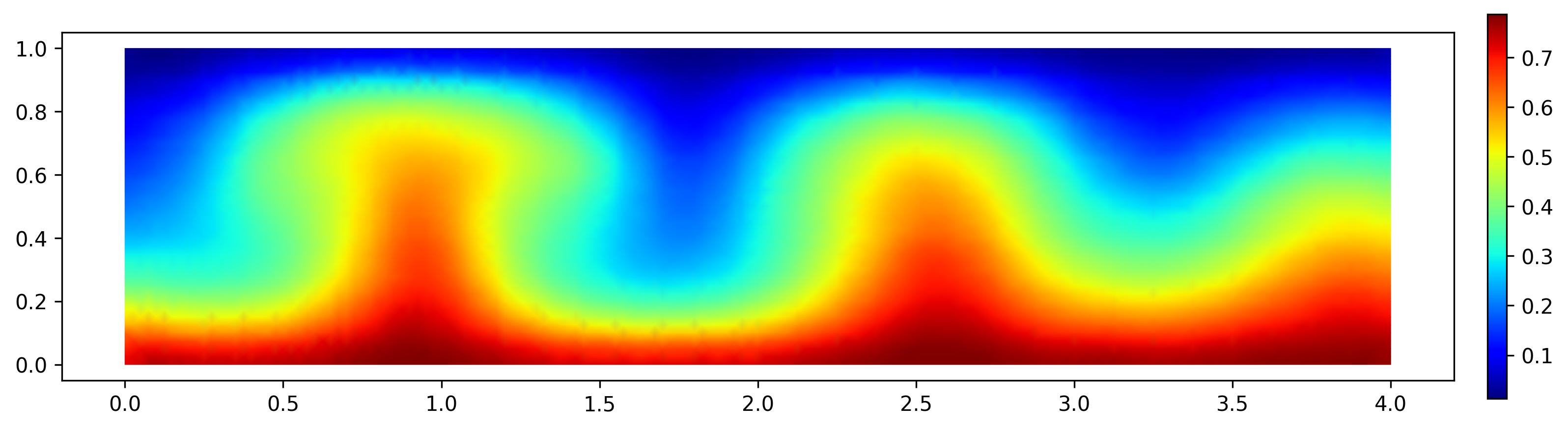}
            \caption{Proposed model}
        \end{subfigure}

    \caption{Contours of temperature as predicted by (a) CFD solver, (b) MeshGraphNets, and (c ) the proposed model in this study for aspect ratio of 4, T-hot =300.8K and T-cold=300K.}
\label{fig:results_diff_temp_300.8}
\end{figure*}

\begin{figure*}[h]

        \centering
        \begin{subfigure}[t]{0.6\textwidth}
        \centering
            \includegraphics[width=8cm, height=2cm]{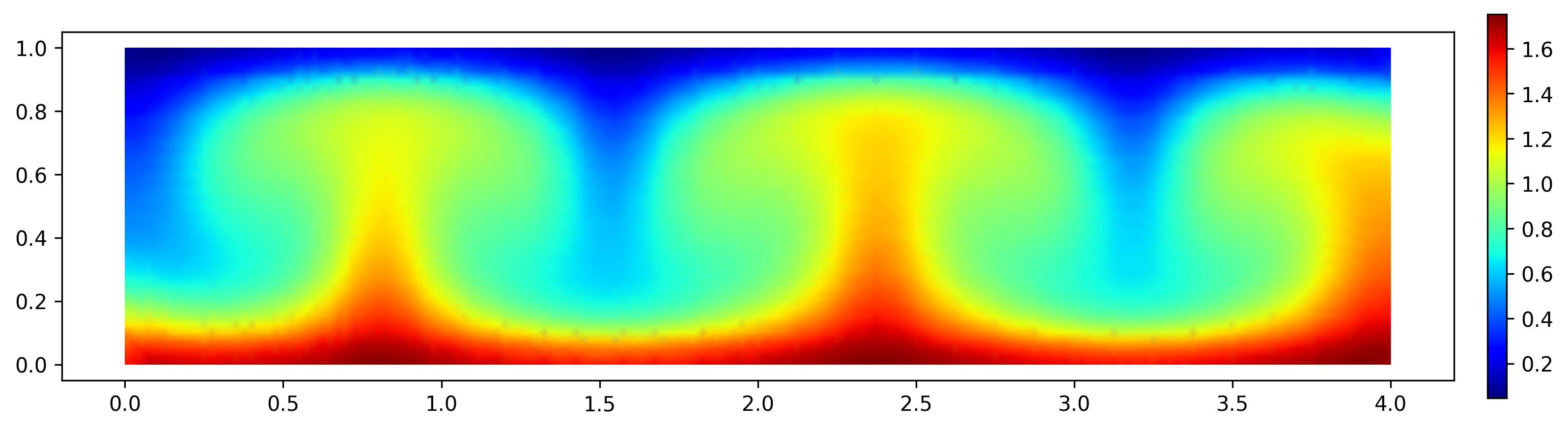}
            \caption{CFD Solver.}
        \end{subfigure}
        \begin{subfigure}[t]{0.6\textwidth}
        \centering
            \includegraphics[width=8cm, height=2cm]{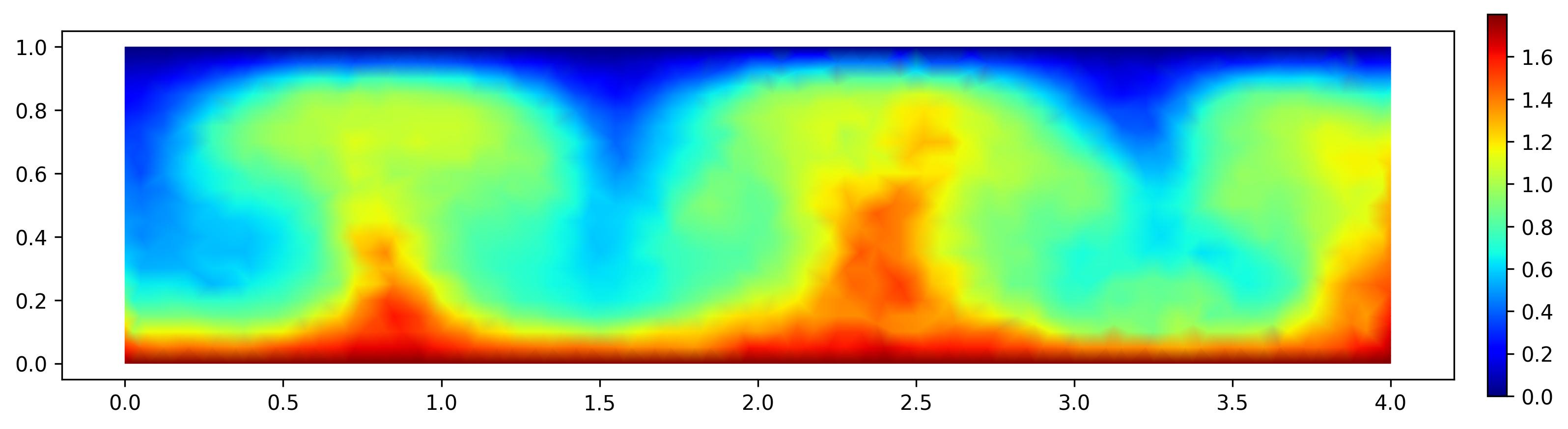}
            \caption{MeshGraphNets}
        \end{subfigure}
        \begin{subfigure}[t]{0.6\textwidth}
        \centering
            \includegraphics[width=8cm, height=2cm]{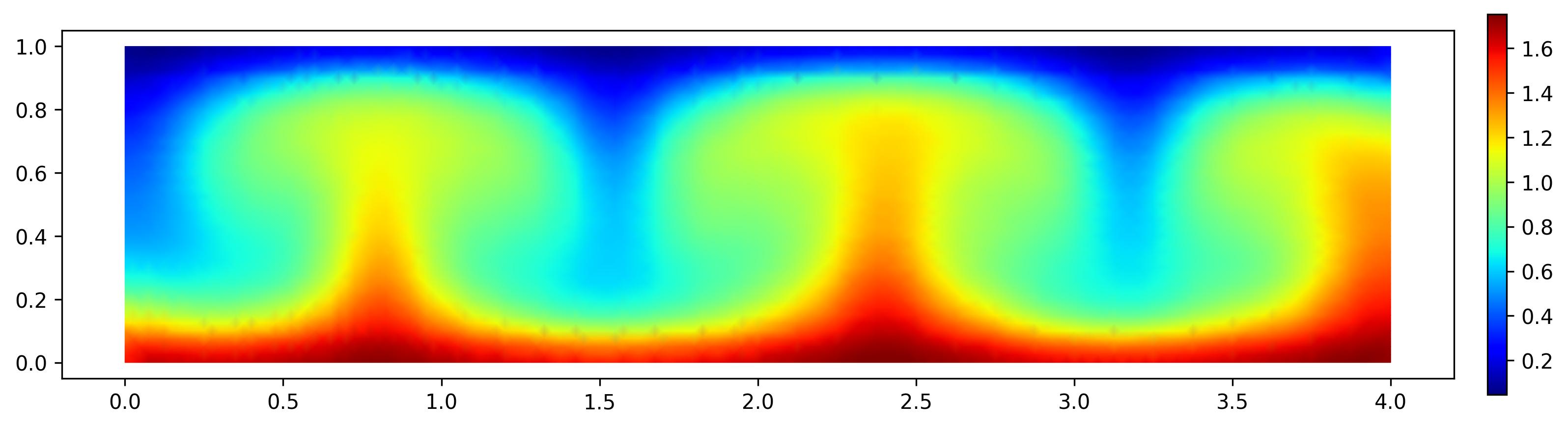}
            \caption{Proposed model}
        \end{subfigure}

    \caption{Contours of temperature as predicted by (a) CFD solver, (b) MeshGraphNets, and (c ) the proposed model in this study for aspect ratio of 4, T-hot =301.8K and T-cold=300K.}
\label{fig:results_diff_temp_301.8}
\end{figure*}

Additional experiments were conducted with the aspect ratio fixed at 1:4 to evaluate model performance under different initial temperature conditions. Fig.~\ref{fig:results_diff_temp_300.8} and Fig.~\ref{fig:results_diff_temp_301.8} represent qualitative results illustrating the flow behavior, highlighting the models’ ability to capture thermal convection patterns. Fig.\ref{fig:results_diff_temp_300.8}, (a)-(c) illustrate a comparison of the predicted temperature contours as obtained by CFD solver, MeshGraphNets~\cite{pfaff2020learning}, and the proposed model in this study, respectively. The aspect ratio of the channel was set to be 1:4, while the bottom and top wall temperatures were fixed at 300.8K and 300K, respectively. A visual comparison demonstrates that the proposed model provides more accurate predictions by closely replicating the ground truth convection patterns. Fig.~\ref{fig:results_diff_temp_301.8} presents a similar experiment, with a different bottom wall temperature which was set to 301.8K and maintained throughout the simulation. As observed, the proposed model yields a higher accuracy of the temprature field when compared to the MeshGraphNets.

%%%%%%%%%%%%%%%%%%%%%%%%%%%%%%%%%%%%%%%%%%%%%
%%%%%%%%%%%%%%%%%%%%%%%%%%%%%%%%%%%%%%%%%%%%%

%%%%%%%%%%%%%%%%%%%%%%%%%%%%%%%%%%%%%%%%%%%%%
%%%%%%%%%%%%%%%%%%%%%%%%%%%%%%%%%%%%%%%%%%%%%

\begin{figure*}
    \centering
    \begin{subfigure}[t]{1\textwidth}

        \centering
        \begin{subfigure}[t]{0.6\textwidth}
        \centering
            \includegraphics[width=8cm, height=2cm]{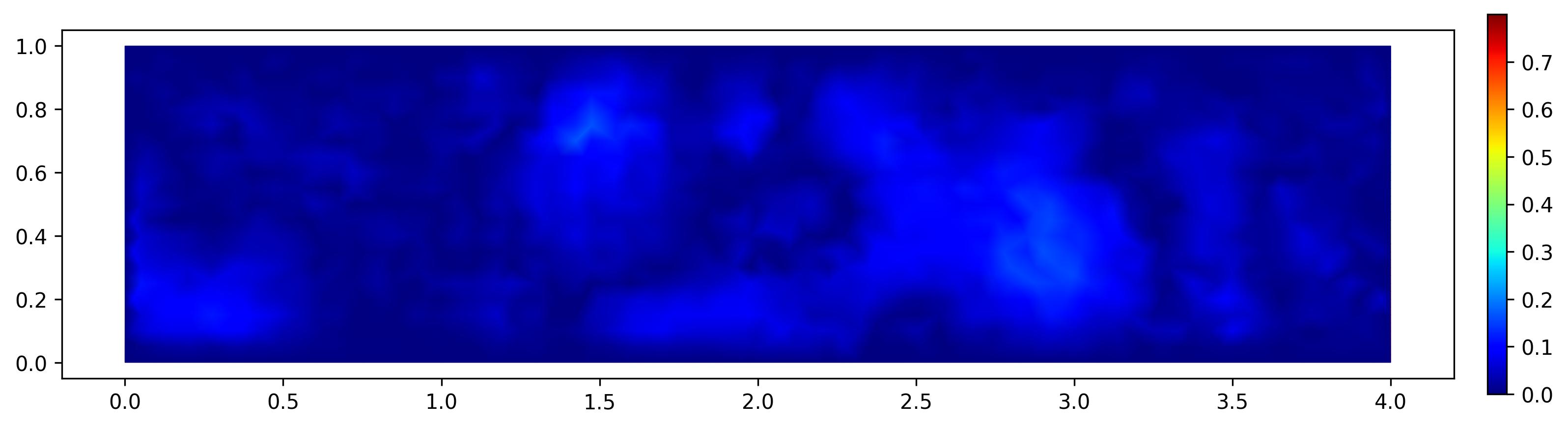}
            \caption{Error map of MeshGraphNets~\cite{pfaff2020learning}.}
        \end{subfigure}

        \centering
        \begin{subfigure}[t]{0.6\textwidth}
        \centering
            \includegraphics[width=8cm, height=2cm]{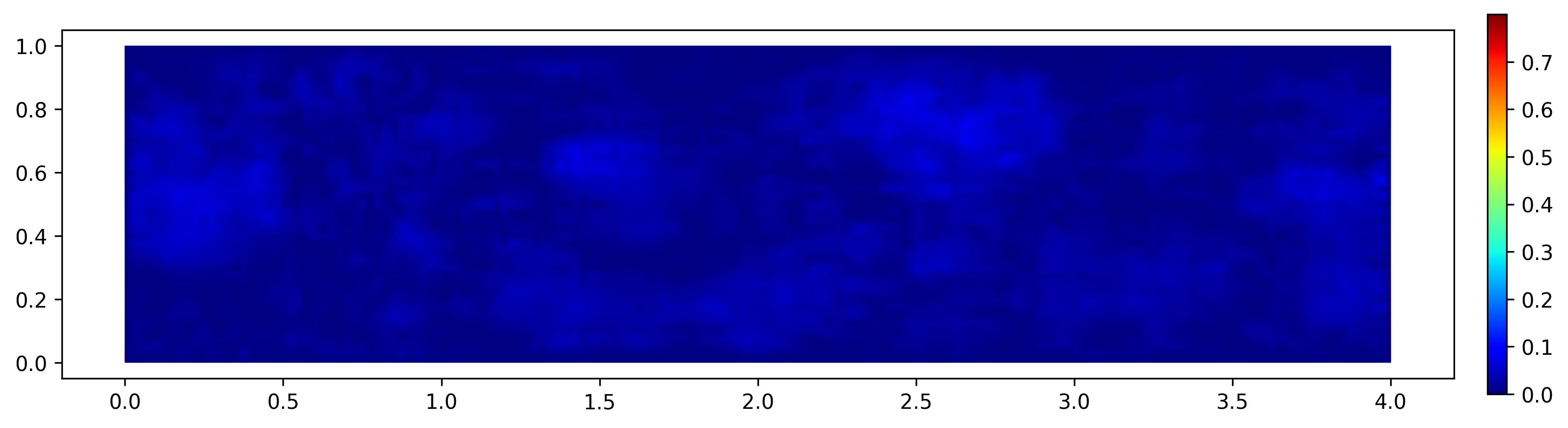}
            \caption{Error map of the proposed model.}
        \end{subfigure}

    \end{subfigure}
    \caption{Temperature prediction error maps for bottom wall temperature of 300.7K, with the CFD solver as the ground truth. (a) Proposed model, (b) MeshGraphNets.}
\label{fig:error_temp_0}
\end{figure*}

\begin{figure*}
    \centering
    \begin{subfigure}[t]{1\textwidth}

        \centering
        \begin{subfigure}[t]{0.6\textwidth}
        \centering
            \includegraphics[width=8cm, height=2cm]{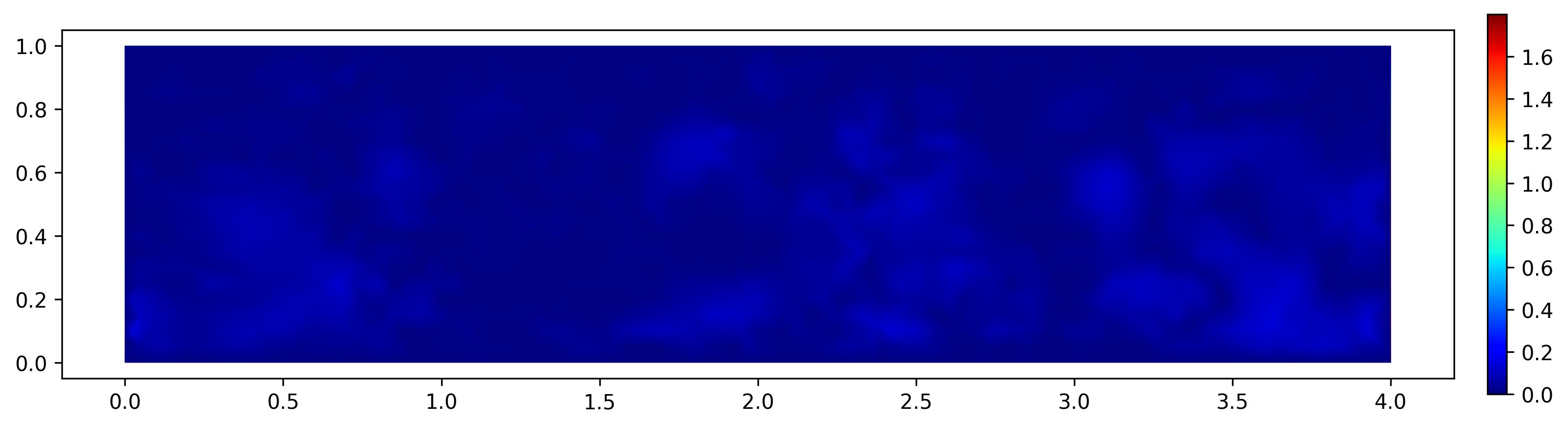}
            \caption{Error map of MeshGraphNets~\cite{pfaff2020learning}.}
        \end{subfigure}

        \centering
        \begin{subfigure}[t]{0.6\textwidth}
        \centering
            \includegraphics[width=8cm, height=2cm]{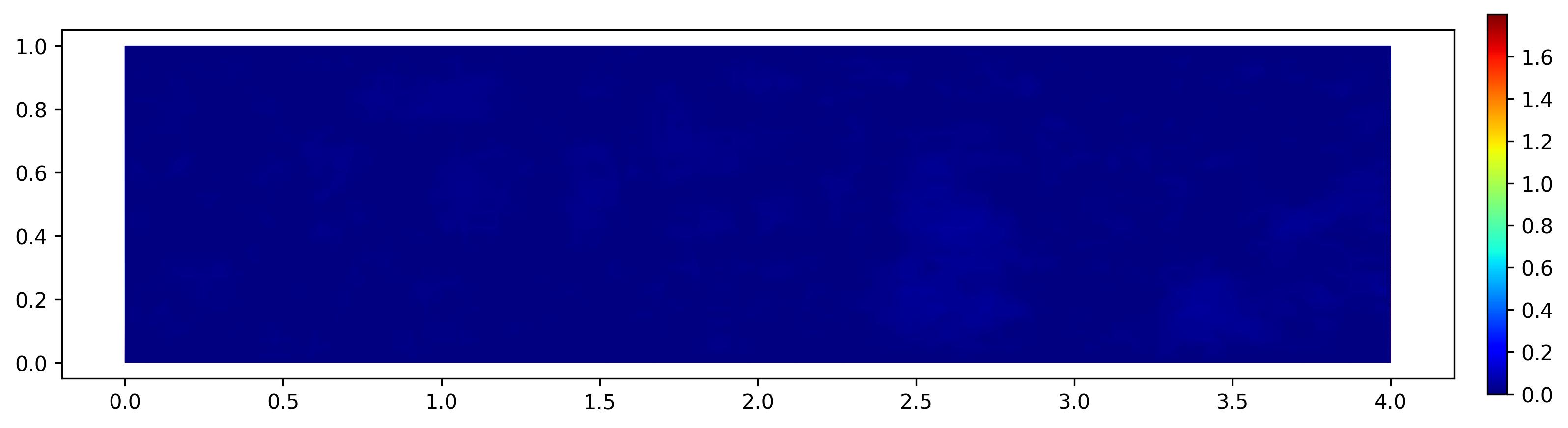}
            \caption{Error map of the proposed model.}
        \end{subfigure}

    \end{subfigure}
    \caption{Temperature prediction error maps for bottom wall temperature of 301.8 K, with the CFD solver as the ground truth. (a) Proposed model, (b) MeshGraphNets.}
\label{fig:error_temp_1}
\end{figure*}

\begin{figure*}
    \centering
    \begin{subfigure}[t]{1\textwidth}
        \centering
        \begin{subfigure}[t]{0.45\textwidth}
            \includegraphics[width=\textwidth]{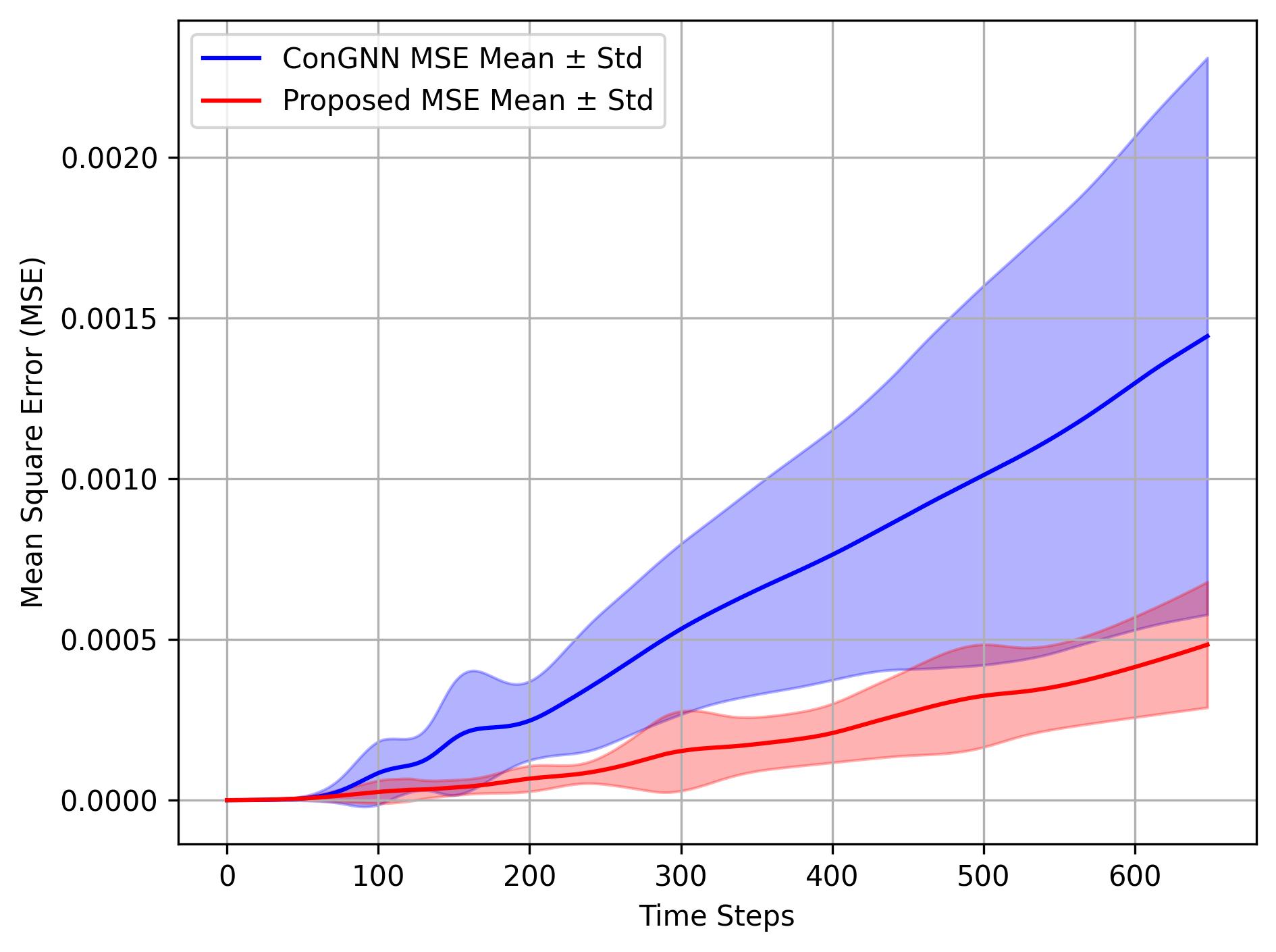}
        \end{subfigure}
        \begin{subfigure}[t]{0.45\textwidth}
            \includegraphics[width=\textwidth]{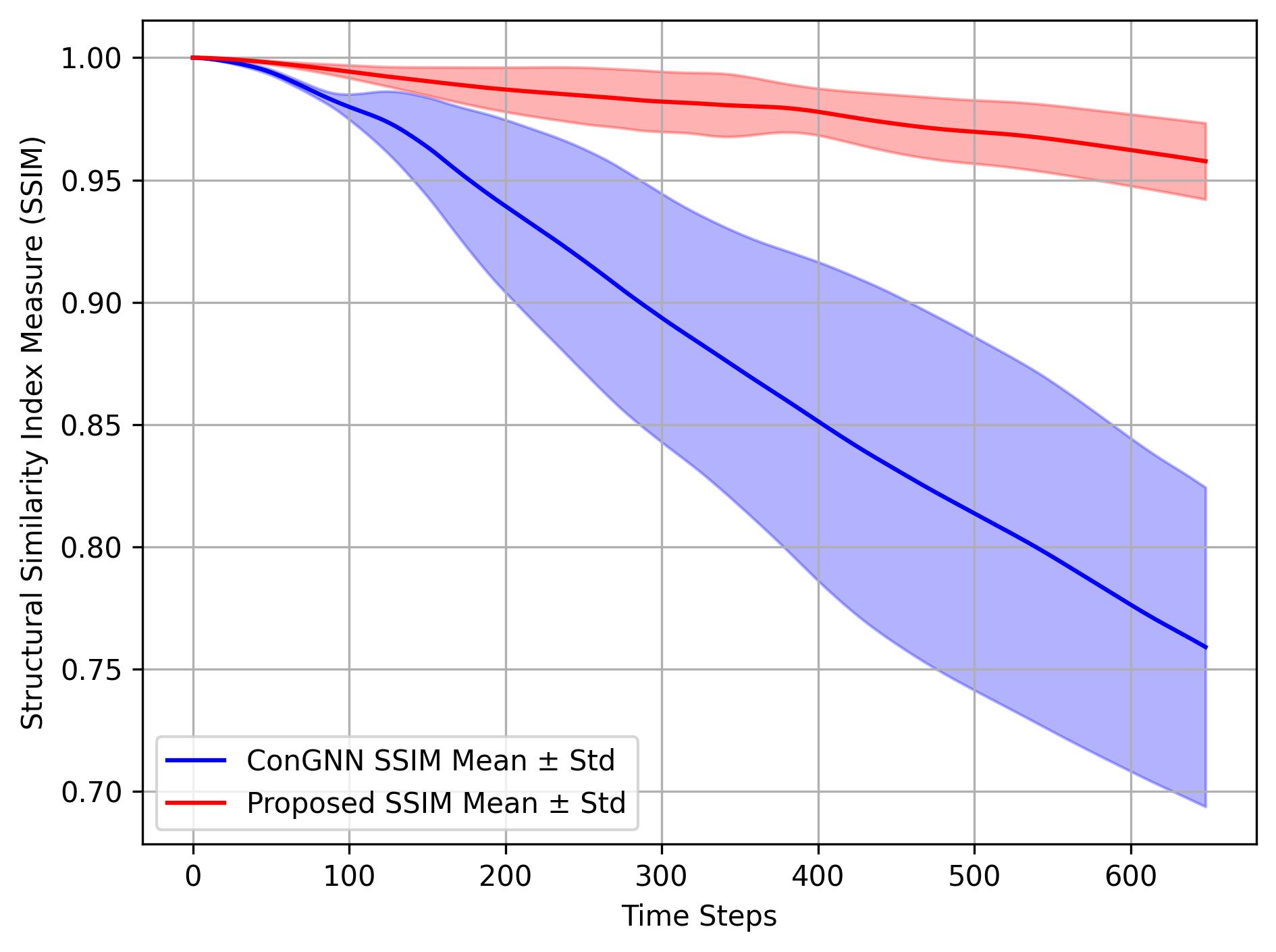}
        \end{subfigure}
    \end{subfigure}
    \caption{Quantitative comparison between the predicted and ground truth temperature fields. The red and blue lines represent the proposed model and MeshGraphNets, respectively. Subfigures (a) and (b) show the average and standard deviation of MSE and SSIM metrics, respectively. In (a), lower MSE values indicate better performance, while in (b), higher SSIM values (closer to 1) reflect better structural similarity and prediction accuracy.}
\label{fig:ssim_mse_diagram}
\end{figure*}

\begin{figure*}
    \centering
    \begin{subfigure}[t]{1\textwidth}
        \centering
        \begin{subfigure}[t]{0.45\textwidth}
            \includegraphics[width=\textwidth]{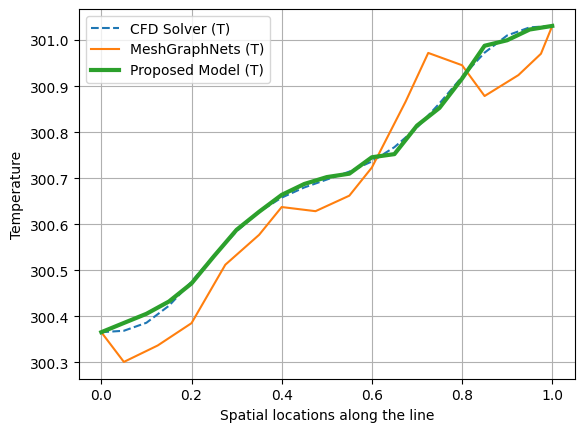}
        \end{subfigure}
        \begin{subfigure}[t]{0.45\textwidth}
            \includegraphics[width=\textwidth]{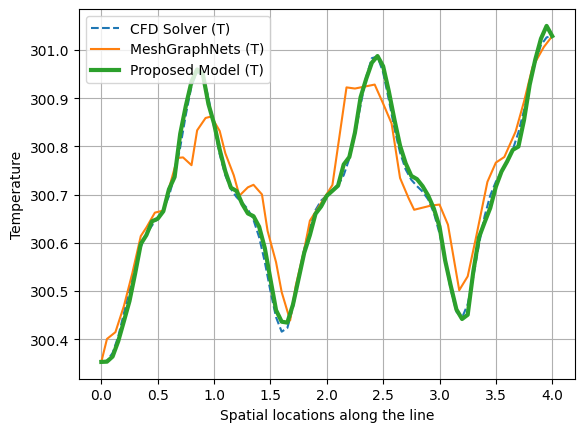}
        \end{subfigure}
    \end{subfigure}
    \caption{Temperature profile comparisons along a horizontal slice at y=0.5 and frame 300, shown with two different aspect ratios. Each subplot includes results from the CFD solver as GT, GraphMeshNets, and the proposed model. (a) The proposed method accurately tracks local variations and peak magnitudes in the temperature profile. (b) The proposed model maintains a consistent and precise match with the ground truth across spatial positions, demonstrating superior spatial fidelity compared to the baseline.}
\label{fig:throughlines}
\end{figure*}

%%%%%%%%%%%%%%%%%%%%%%%%%%%%%%%%%%%%%%%%%%%%%
%%%%%%%%%%%%%%%%%%%%%%%%%%%%%%%%%%%%%%%%%%%%%

Figs.~\ref{fig:error_temp_0} and \ref{fig:error_temp_1} present the error maps at time step 300 for MeshGraphNets and the proposed model, for the bottom wall temperatures of 300.8 K and 301.8 K, respectively. These maps illustrate the spatial distribution of prediction errors across the enclosure. The error is computed as the absolute difference between the CFD solver output and the corresponding predicted frame. As shown, the proposed model consistently outperforms MeshGraphNets, yielding markedly lower accumulated errors and better alignment with the ground truth thermal distribution.

We further conduct a temporal analysis of the Mean Squared Error (MSE) and Structural Similarity Index Measure (SSIM) between the predicted and ground truth temperature fields for both the proposed model and MeshGraphNets across all time steps. Fig.~\ref{fig:ssim_mse_diagram} illustrates these results, where red and blue lines represent the proposed model and MeshGraphNets, respectively. Fig.~\ref{fig:ssim_mse_diagram}-a and Fig.~\ref{fig:ssim_mse_diagram}-b present the average and standard deviation of MSE and SSIM over time. MSE quantifies pixel-wise differences between predicted and reference fields, with lower values indicating better accuracy, whereas SSIM evaluates perceptual similarity by considering luminance, contrast, and structure, with values closer to 1 reflecting stronger correspondence. As shown, MeshGraphNets exhibits increasing error and decreasing SSIM over time, indicating degraded performance in long-term predictions. In contrast, the proposed model maintains lower MSE and higher SSIM throughout the entire sequence, demonstrating its ability to preserve temporal dynamics more faithfully.

Fig.\ref{fig:throughlines} presents a comparative visualization of temperature distributions along a horizontal line at y=0.5 at time step 300 for aspect ratios of 1:3 and 1:4. Each subplot includes the CFD solver as GT, MeshGraphNets, and the proposed model. As observed in Fig.~\ref{fig:throughlines} (a), the proposed model closely follows the ground truth temperature profile, effectively capturing local variations and peak magnitudes. Similarly, Fig.~\ref{fig:throughlines}(b) shows that the proposed model maintains an accurate match with the ground truth across spatial positions. These results highlight the improved spatial fidelity and predictive performance of the proposed model compared with MeshGraphNets.

%%%%%%%%%%%%%%%%%%%%%%%%%%%%%%%%%%%%%%%%%%%%%%%%%%%

\subsubsection{Training Efficiency and Computational Performance}
\label{Computational}

We evaluate training efficiency relative to MeshGraphNets under identical rollout horizons, loss functions, and batch sizes. On a lower-resolution case ($\sim1,750$ nodes), training MGN to convergence required $\sim24$ hours. In contrast, our multi-stage model reached comparable or better validation error in 5-6 hours-an $\approx4×$ speedup. All runs were performed on a laptop (Intel Core i9, 16 GB RAM, NVIDIA RTX 3080 Ti, Alienware). This gain reflects both the reduced per-step cost at coarse scales and more stable optimization due to our staged curriculum, underscoring the practicality and scalability of the proposed approach.

\section{Conclusion}
\label{Conclusion}

We addressed the challenge of learning fluid-thermal dynamics on high-resolution structured quadrilateral meshes by introducing a multi-stage GNN tailored to heat-transfer prediction. While traditional CFD delivers high fidelity at significant computational cost, and single-scale GNN surrogates often struggle on fine meshes, our architecture couples hierarchical pooling/unpooling with parallel message-passing branches to capture near-wall thermal/velocity gradients and long-range buoyant couplings simultaneously.
We validated the method on a new CFD dataset of natural convection in rectangular cavities with adiabatic sidewalls, hot bottom, and cold top, spanning multiple aspect ratios and exhibiting Rayleigh-Bénard cellular patterns. Across metrics, the proposed model outperforms MGN and other strong GNN baselines, delivering higher predictive accuracy, faster training, and substantially lower error drift in long autoregressive rollouts. By improving convergence stability and reducing cumulative rollout error, our framework offers a scalable, efficient surrogate for mesh-based heat-transfer simulations in enclosure flows-reducing reliance on repeated high-fidelity solver runs while preserving physically meaningful structure.

\bibliography{main}

%%%%%%%%%%%%%%%%%%%%%%%%%%%%%%%%%%%%%%%%%%%%%%%%%%%%%%%%%%%%%%%%%%%%
\newpage
\appendix
\section*{Appendix:}
\label{appendix}

\subsection{Fast pooling and un-pooling implementation:}
\label{appendix:pooling_unpoolin}
The original implementations of the pooling and unpooling operators are computationally expensive, which considerably slows down the training process and hinders overall model efficiency~\cite{deshpande2022magnet}. To address this limitation, we introduce an optimized implementation of the pooling and unpooling operations that significantly reduces computational overhead.

In the pooling, we down-sample the input graph by aggregating features from clustered nodes, where each cluster forms a node in the pooled graph. The input tensor x has a shape of [B, C * N], representing B batches of node features with C channels for each of the N original nodes. To define clusters, we use a fixed-size subgraph index matrix of shape [n cliques, 3], where each row contains the indices of nodes grouped into a cliques. Since some cliques may contain fewer than three nodes, we duplicate indices to maintain a consistent size of three per row. This allows us to use efficient tensor operations (instead of for-loops which is slower) to extract and concatenate node features across the clusters, and then compute the average to obtain a pooled representation. This approach significantly improves computational efficiency while maintaining flexibility for variable-size clusters. The pooling implementation is presented in Listing.~\ref{lst:pooling_implementation}.

The Unpooling module reconstructs a high-resolution node feature map from a down-sampled graph representation by distributing pooled features back to their corresponding original nodes. The input tensor x has shape [B, C * M], where B is the batch size, C is the number of feature channels per clique, and M is the number of cliques. Each clique corresponds to a set of original nodes, whose indices are stored in \textit{self.subgraph}, a list where each element contains the node indices belonging to a clique. To efficiently map features back, the tensor is reshaped and broadcasted such that each cluster's feature vector is assigned to its corresponding original nodes using direct indexing, again avoiding slower for-loops. As clusters may overlap (i.e., a node may appear in multiple clusters), the same feature is written multiple times; depending on the application, one could also apply averaging or summation later. This matrix-based ungrouping is both efficient and scalable for restoring the graph structure after the grouping. The un-pooling implementation is presented in Listing.~\ref{lst:unpooling_implementation}.\\

\begin{lstlisting}[style=pythonstyle, 
caption={Graph Pooling Module (G\_Pool)}, 
basicstyle=\tiny\ttfamily,
label={lst:pooling_implementation}]
class G_Pool(nn.Module):
    def __init__(self, subgraph, nodes=None):
        super(G_Pool, self).__init__()
        self.subgraph = subgraph
        self.subgraph_tensor = torch.tensor(self.subgraph, dtype=torch.long)  # shape [2624, 3]
        self.nodes = nodes

    def forward(self, x):
        batch_size = x.size(0)
        total_units = x.size(1)
        n_channels = total_units // self.nodes
        n_cliques = len(self.subgraph)
        x = x.view(batch_size, n_channels, self.nodes)  # (B, C, N)

        x_reshaped = Rearrange(x, 'b c n -> (b n) c')
        
        x_selected_1 = x_reshaped[self.subgraph[:,0],:].unsqueeze(2)
        x_selected_2 = x_reshaped[self.subgraph[:,1],:].unsqueeze(2)
        x_selected_3 = x_reshaped[self.subgraph[:,2],:].unsqueeze(2)

        x_selected_1_2_3 = torch.cat((x_selected_1, x_selected_2, x_selected_3), dim=2)
        pooled_out = torch.mean(x_selected_1_2_3, dim=2)

        pooled_out = Rearrange(pooled_out, ' (b n) c -> b (c n) ', b=batch_size)
        return pooled_out  # shape: (B, C * n_cliques)

\end{lstlisting}

\begin{lstlisting}[style=pythonstyle, 
caption={Graph Unpooling Module (G\_Unpool)}, 
basicstyle=\tiny\ttfamily,
label={lst:unpooling_implementation}]]
class G_Unpool(nn.Module):
    def __init__(self, subgraph, nodes=None):
        super(G_Unpool, self).__init__()
        self.device = torch.device('cuda' if torch.cuda.is_available() else 'cpu')
        self.nodes = nodes
        self.subgraph = [torch.tensor(sg, dtype=torch.long) for sg in subgraph]

    def forward(self, x):
        B, total_units = x.shape
        M = len(self.subgraph)
        assert self.nodes is not None and total_units % M == 0, "Invalid input or missing 'nodes'."
        C = total_units // M
        x = x.view(B, C, M)  # (B, C, M)
        x = Rearrange(x, 'b c n -> (b n) c')

        device = x.device
        self.subgraph = torch.stack(self.subgraph)
        unpooled = torch.zeros(self.nodes, B * C).to(device)

        unpooled[self.subgraph[:,0]] = x
        unpooled[self.subgraph[:,1]] = x
        unpooled[self.subgraph[:,2]] = x

        unpooled = Rearrange(unpooled, 'n (b c) -> b (n c)', c=C)
        return unpooled.view(B, C * self.nodes)
\end{lstlisting}

%%%%%%%%%%%%%%%%%%%%%%%%%%%%%%%%%%%%%%%%%%%%%%%%%%%%%%%%%%%%%%%%%%%%

\end{document}